\documentclass[10pt,journal,compsoc]{IEEEtran}
\ifCLASSOPTIONcompsoc
  \usepackage[nocompress]{cite}
\else
  \usepackage{cite}
\fi
\ifCLASSINFOpdf
\else
\fi
\usepackage{amsmath}
\usepackage{array}

\usepackage[caption=false,font=footnotesize]{subfig}
\usepackage{url}

\usepackage{graphicx}
\usepackage{booktabs}
\usepackage{multirow}
\usepackage{float}
\usepackage{dsfont}
\usepackage{amssymb}
\usepackage[boxed,vlined]{algorithm2e}
\usepackage{cite}
\usepackage{xcolor}
\usepackage{threeparttable}
\usepackage{amsopn}

\newcommand{\etal}{\mbox{\emph{et al.\ }}}

\newcommand{\ie}{\mbox{\emph{i.e.,\ }}}
\newcommand{\eg}{\mbox{\emph{e.g.,\ }}}
\newcommand{\tabincell}[2]{\begin{tabular}{@{}#1@{}}#2\end{tabular}}

\begin{document}
%
\title{Tattoo Image Search at Scale: Joint Detection and Compact Representation Learning}
%
%
%
%

\author{Hu~Han,~\IEEEmembership{Member,~IEEE,}
        Jie~Li,
        Anil~K.~Jain,~\IEEEmembership{Fellow,~IEEE,}\\
        Shiguang~Shan,~\IEEEmembership{Senior Member,~IEEE}
        and~Xilin~Chen,~\IEEEmembership{Fellow,~IEEE}
\IEEEcompsocitemizethanks{\IEEEcompsocthanksitem Hu Han, Jie Li, Shiguang Shan, and Xilin Chen are with the Key Laboratory of Intelligent Information Processing of Chinese Academy of Sciences (CAS), Institute of Computing Technology, CAS, Beijing, 100190, China \protect\\
Anil K. Jain is with the Department of Computer Science and Engineering, Michigan State University, East Lansing, MI 48824, USA.\protect\\
E-mail: \{hanhu, sgshan, xlchen\}@ict.ac.cn; jain@cse.msu.edu; jie.li@vipl.ict.cn}
}


%
%

\markboth{}%
{Han \MakeLowercase{\textit{et al.}}: Tattoo Image Search at Scale: Joint Detection and Compact Representation Learning}
%



\IEEEtitleabstractindextext{%
\begin{abstract}

The explosive growth of digital images in video surveillance and social media has led to the significant need for efficient search of persons of interest in law enforcement and forensic applications.
Despite tremendous progress in primary biometric traits (\eg face and fingerprint) based person identification, a single biometric trait alone can not meet the desired recognition accuracy in forensic scenarios.
Tattoos, as one of the important soft biometric traits, have been found to be valuable for assisting in person identification.
However, tattoo search in a large collection of unconstrained images remains a difficult problem, and existing tattoo search methods mainly focus on matching cropped tattoos, which is different from real application scenarios.
To close the gap, we propose an efficient tattoo search approach that is able to learn tattoo detection and compact representation jointly in a single convolutional neural network (CNN) via multi-task learning.
While the features in the backbone network are shared by both tattoo detection and compact representation learning, individual latent layers of each sub-network optimize the shared features toward the detection and feature learning tasks, respectively.
We resolve the small batch size issue inside the joint tattoo detection and compact representation learning network via random image stitch and preceding feature buffering.
We evaluate the proposed tattoo search system using multiple public-domain tattoo benchmarks, and a gallery set with about $300$K distracter tattoo images compiled from these datasets and images from the Internet.
In addition, we also introduce a tattoo sketch dataset containing $300$ tattoos for sketch-based tattoo search.
Experimental results show that the proposed approach has superior performance in tattoo detection and tattoo search at scale compared to several state-of-the-art tattoo retrieval algorithms.

\end{abstract}

}

\maketitle

\IEEEdisplaynontitleabstractindextext

%
\IEEEpeerreviewmaketitle

\IEEEraisesectionheading{\section{Introduction}\label{sec:introduction}}

%
%
%
%

\IEEEPARstart{I}{n} the past few decades, because of the advances in computing, imaging, and Internet technologies, digital images and videos are now widely used for representing information in video surveillance, and social media.
In 2017, IHS Markit estimated that the United States has approximately $50$ million surveillance cameras, and China has about $176$ million.\footnote{\url{http://www.straitstimes.com/opinion/chinas-all-seeing-surveillance-state-is-reading-its-citizens-faces}}
Another statistics by YouTube in 2017 shows that the length of videos uploaded to YouTube every minute is approximately $300$ hours.\footnote{\url{https://www.statisticbrain.com/youtube-statistics}}
Given the explosive growth of image and video data, there is great demand for efficient instance search technologies, particularly for persons of interest in law enforcement and forensics.
Although tremendous progress has been made in face recognition based person identification, many situations exist where face recognition cannot identify an individual with sufficiently high accuracy.
This is especially true when the face image quality is poor or the persons of interest intentionally hide their faces.
\begin{figure}[t]
\centering
\includegraphics[width=0.6\linewidth]{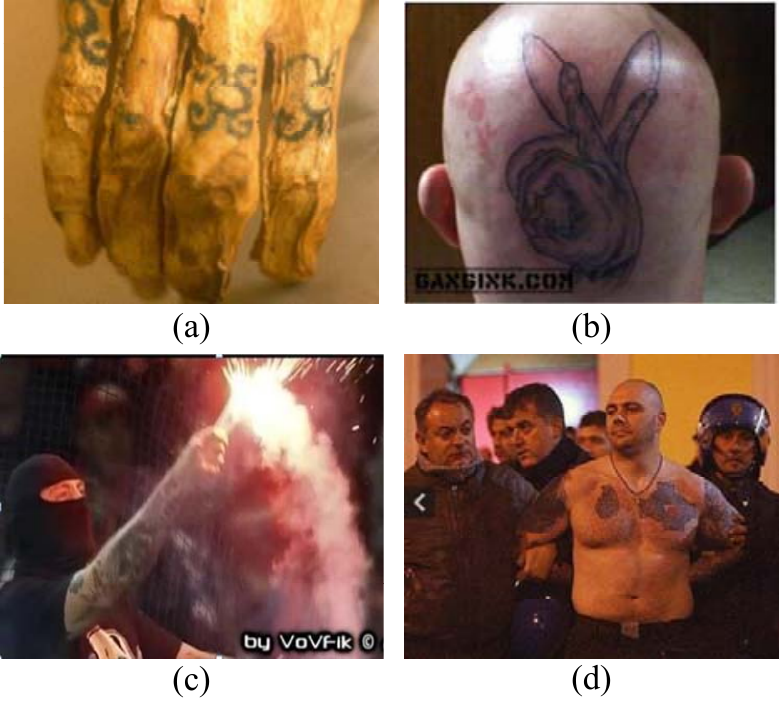}
\vspace{-4mm}
\caption[TattooExamples]{Tattoo examples: (a) a tattoo on the right hand of a Chiribaya mummy in southern Peru who lived from A.D. 900 to 1350,\footnotemark[4] (b) a tattoo on the head signifying gang membership association,\footnotemark[5] (c, d) tattoos of a masked ringleader of the riots during Euro 2012 qualifier, and the suspect of the masked ringleader identified by tattoos.\footnotemark[6]}
\label{Fig.TattooExamples}
\vspace{-6mm}
\end{figure}
In such cases, it is critical to acquire supplementary information to assist in person identification \cite{LeeMM12}.
On the basis of this rationale, since 2009 the US Federal Bureau of Investigation (FBI) has been working on extending the capability of its Integrated Automated Fingerprint Identification System (IAFIS) by using additional biometric modalities, including iris, palm print, scars, marks, and tattoos\footnote{Scars, marks, and tattoos are collectively referred to as SMT.}.
This new person identification system is named the Next Generation Identification (NGI) system \cite{NGI10}, which is able to offer state-of-the-art biometric identification services for homeland security, law enforcement, etc.

\footnotetext[4]{\url{https://www.smithsonianmag.com/history/tattoos-144038580}}
\footnotetext[5]{\url{http://www.gangink.com/index.php?pr=GANG_LIST}}
\footnotetext[6]{\url{http://www.telegraph.co.uk/sport/football/teams/serbia/8061619/Masked-ringleader-of-crowd-trouble-during-Italy-Serbia-clash-identified-by-tattoos.html}}

Among the various soft biometric traits, tattoos, in particular, have received substantial attention over the past several
years due to their prevalence among the criminal section of the population and their saliency in visual attention.
Humans have marked their bodies with tattoos to express personal beliefs or to signify group association for more than $5,000$ years (see Fig. \ref{Fig.TattooExamples} (a)).
Figs. \ref{Fig.TattooExamples} (c, d) show an example how a suspect of the masked ringleader of the riots during Euro 2012 qualifier was arrested based on the tattoos on his arms.
In fact, criminal investigations have leveraged soft biometric traits as far back as the late $19$th century \cite{Bertillon96}.
For example, the first personal identification system, the Bertillon system, tried to provide a precise and scientific method to identify criminals by using physical measurements of body parts, especially measurements of the head and face, as well as the images of SMT on the body.
Tattoos were also reported to be useful for assisting in identifying victims of terrorist attacks such as $9/11$ and natural disasters like the $2004$ Indian Ocean tsunami \cite{Jain07}.
Nowadays, law enforcement agencies in the US routinely photograph and catalog tattoo patterns for use in identifying victims and convicts.
The NIST Tatt-C and Tatt-E challenges have been helpful in advancing the development of tattoo detection and identification systems for real application scenarios \cite{NganNIST15}.

Despite the value of tattoos for assisting in person identification, putting it to practical use has been difficult.
Unlike primary biometric traits, much variability exists in pattern types of tattoos.
The early use of tattoo for assisting in person identification relied heavily on manual annotations and comparisons.
This has motivated the study of automatic identification algorithms for tattoos \cite{McCabeTR00,Jain07,LeeCVPR08,ActonSSIAI08,MangerCRV12,HanICB13,NganNIST15,XuICB16}.

Despite the progress in tattoo retrieval, existing methods have some serious limitations.
In fact, most of the current practice of tattoo matching aims at tattoo identification, and not learning a compact representation for efficient tattoo search at scale.
More importantly, existing tattoo identification methods primarily focus on matching cropped tattoos, which does not replicate the in-situ scenarios, where the search must be operated in raw images or video frames.
In addition, while sketch to photo matching has been widely studied in the areas such as image retrieval \cite{CaoCVPR11,CaoACMMM10} and face recognition \cite{WangTPAMI09,HanTIFS13}, research on sketch based tattoo search is very limited \cite{HanICB13}.

\subsection{Proposed Approach}
\label{Sec.ProposedApproach}

\begin{figure}[t]
\centering
\includegraphics[width=0.85\linewidth,height=0.7\linewidth]{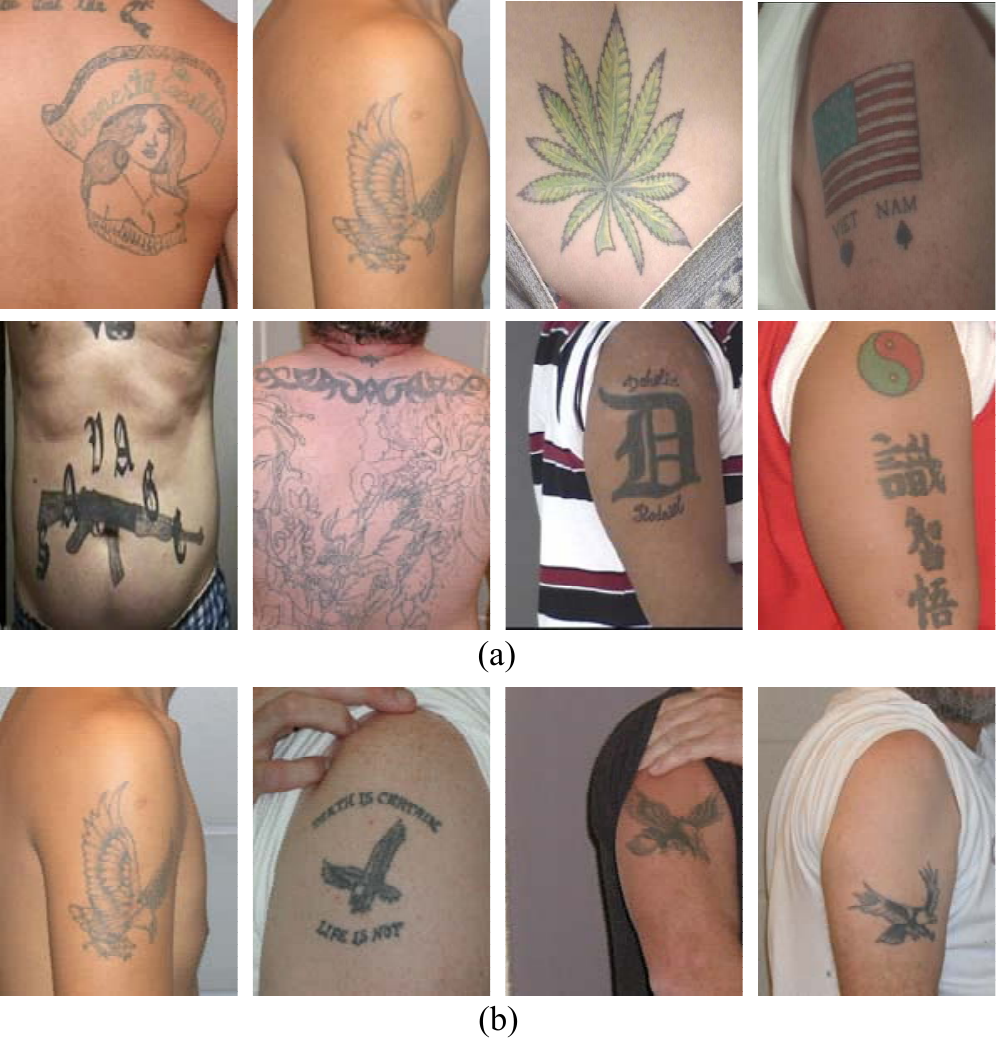}
\vspace{-4mm}
\caption[LargeDiversity]{Tattoo images from the Tatt-C dataset \cite{NganNIST15} suggest that tattoo search is challenging because (a) there are various types of tattoos containing categories such as humans, animals, plants, flags, objects, abstract, symbols, etc., and (b) the intra-class variability in one class of tattoos (\ie eagle) can be very large.}
\label{Fig.LargeDiversity}
\vspace{-4mm}
\end{figure}

To overcome the above limitations of the current tattoo search methods, we present a joint detection and compact representation learning approach for tattoo search at scale.
The proposed approach is motivated by recent advances in object detection and compact representation learning, but takes into account the unique challenges in a tattoo search domain, such as large intra-class variability, poor image quality, and image deformations (see Fig. \ref{Fig.LargeDiversity}).
In addition, the proposed approach can be trained in a fully end-to-end fashion, and can leverage additional operational data to improve the tattoo search.

As shown in Fig. \ref{Fig.JointDetFL}, the proposed approach handles tattoo detection and compact representation learning in a single convolutional neural network (CNN) via multi-task learning.
Given an input image, a shared feature map is first computed via a deep CNN network, which is then fed into individual sub-networks, which aim at tattoo detection and compact feature learning tasks, respectively.

The main contributions of this paper include: (i) the first end-to-end trainable approach for joint tattoo detection and compact representation learning in a single network allowing more robust and discriminative feature learning via feature sharing; (ii) effective strategies in resolving small batch size issue w.r.t. the compact representation learning module of the network; (iii) superior performance and much lower computational cost compared to the state-of-the-art algorithms; and (iv) compiling a dataset with $300,000$ tattoo images in the wild and thousands of annotations for large-scale tattoo search, and a dataset with $300$ tattoo sketches (see Fig. \ref{Fig.TattooSearchExamples} (b)) for sketch-based tattoo search; both datasets will be put into the public domain.

Our preliminary work of this research is described in \cite{HanICB13}.
Essential improvements over \cite{HanICB13} include: (i) use of task-driven learned features instead of hand-crafted features for joint tattoo detection and compact representation learning in a single network; (ii) extensions to large-scale tattoo search via compact feature learning; and (iii) compiling a sketch dataset for studying sketch-based tattoo search.

The remainder of this paper is structured as follows.
We briefly review related literature in Section~\ref{Sec.RelatedWork}.
The details of the proposed tattoo search approach are provided in Section~\ref{Sec.ProposedApproach}.
In Section~\ref{Sec.Experiment}, we introduce the WebTattoo and tattoo sketch datasets, and provide the experimental results and analysis.
Finally, we conclude this work in Section~\ref{Sec.Summary}.

\section{Related Work}
\label{Sec.RelatedWork}

\subsection{Tattoo Identification and Retrieval}
\label{Sec.TattooRetrieval}

In the following, we briefly review the literature on tattoo identification and retrieval, covering detection, feature representation, databases, and performance (see Table \ref{tab:TattooRetrievalMethods}).

The early practice of tattoo image retrieval relied on keywords or metadata based matching.
For example, law enforcement agencies usually follow the ANSI/NIST-ITL 1-2000 standard \cite{McCabeTR00} for assigning a single keyword to each tattoo image in the database.
However, a keyword-based tattoo image retrieval has several limitations in practice \cite{LeeMM12}: (i) The classes defined by ANSI/NIST-ITL offer a limited vocabulary which is insufficient for describing various tattoo patterns; (ii) multiple keywords may be needed to adequately describe a tattoo image; (iii) human annotation is subjective and different subjects can give dramatically different labels to the same tattoo image.

\begin{table*}[!ht]
\begin{center}
\caption{A summary of published methods on tattoo identification and retrieval.}
\label{tab:TattooRetrievalMethods}
\vspace{-1mm}
\scriptsize
\begin{tabular}{lllll}
\toprule
\textbf{Publication} & \textbf{Detection model} & \tabincell{c}{\textbf{Feature and retrieval model}} & \tabincell{c}{\textbf{Tattoo database} \\ \#images (query; target)} & \tabincell{c}{\textbf{Results}} \\
\hline

\tabincell{c}{Jain \etal \cite{Jain07,LeeMM12}\\
(2007,2012)} & \tabincell{l}{Gradient thresholding} & \tabincell{l}{Color histogram and correlogram; \\ shape moments;\\ edge direction coherence; \\ Fusion of per feature similarities} & \tabincell{l}{Tattoos from web\\ ($2,157$; $43,140$)$^1$} & \tabincell{l}{46\% prec.@60\% recall} \\
\cline{1-5}


\tabincell{c}{Acton and Rossi \cite{ActonSSIAI08}\\(2008)} & \tabincell{l}{Active contour \\segmentation and \\skin detection} & \tabincell{l}{Global and local features of \\ edge and color;\\ Vector-wise Euclidean distance} & \tabincell{l}{Recreational ($30$; $\sim4,000$)$^2$; \\ Gang ($39$; $\sim4,000$)$^2$} & \tabincell{l}{Recreational: 94.7\% acc.@rank-1; \\ Gang: 82.2\% acc.@rank-1} \\
\cline{1-5}

\tabincell{c}{Jain \etal \cite{JainICIP09}\\(2009)} & \tabincell{l}{Pre-cropped tattoos} & \tabincell{l}{SIFT features with geometric constraint;\\ indexing with location and keyword;\\ Keypoint-wise matching} & \tabincell{l}{MSP ($1,000$; $63,592$)} & \tabincell{l}{85.9\% acc.@rank-1} \\
\cline{1-5}
\tabincell{c}{Li \etal \cite{LiMMRM09}\\(2009)} & n/a & \tabincell{l}{SIFT features;\\ Bag-of-words; Re-ranking} & \tabincell{l}{MSP and ESP\\ ($995$; $101,754$)$^3$} & \tabincell{l}{67\% acc.@rank-1} \\
\cline{1-5}

\tabincell{c}{D. Manger \cite{MangerCRV12}\\(2012)} & n/a & \tabincell{l}{SIFT features;\\ Bag-of-words, hamming embedding,\\ and weak geometry consistency} & \tabincell{l}{German police \\($417$; $327,049$)} & \tabincell{l}{78\% acc.@rank-1} \\
\cline{1-5}

\tabincell{c}{Heflin \etal \cite{HeflinBTAS12}\\(2012)} & \tabincell{l}{Automatic GrabCut \\ and quasi connected \\components} & \tabincell{l}{LBP-like features, SVM} & \tabincell{l}{Tattoo classification\\ ($50$; $500$)$^4$} & \tabincell{l}{85\% acc.@10\% FAR on average,\\ for 15 classes}\\
\cline{1-5}

\tabincell{c}{Han and Jain \cite{HanICB13}\\(2013)} & \tabincell{l}{Pre-cropped tattoos} & \tabincell{l}{SIFT features;\\ Sparse representation classification} & \tabincell{l}{MSU Sketch Tattoo\\ ($100$; $10,100$)} & \tabincell{l}{48\% acc.@rank-100} \\
\cline{1-5}

\tabincell{c}{Wilber \etal \cite{WilberWACV14}\\(2014)} & \tabincell{l}{Pre-cropped tattoos} & \tabincell{l}{Exemplar code using HoG features;\\ Random forest classifier} & \tabincell{l}{$238$ tattoos of 5 classes} & \tabincell{l}{63.8\% avg. acc. for 5 classes} \\
\cline{1-5}

\tabincell{c}{Xu \etal \cite{XuICPR16}\\(2016)} & \tabincell{l}{Skin segmentation\\ and block based\\ decision tree} & \tabincell{l}{Boundary features; \\Shape matching via \\coherent point drift} & \tabincell{l}{Full body tattoo sketch\\ ($547$; $1,641$)} & \tabincell{l}{52.38\% acc.@rank-50} \\
\cline{1-5}

\tabincell{c}{Kim \etal \cite{KimHST16}\\(2016)} & \tabincell{l}{Graphcut} & \tabincell{l}{n/a} & \tabincell{l}{Tatt-C (Detection):  \\$6,308$ images; \\ Evil (Detection): \\$1,105$ images} & \tabincell{l}{Tatt-C: 70.5\% acc.@41\%recall\\ \\ Evil: 69.9\% acc.@67.0\%recall} \\
\cline{1-5}

\tabincell{c}{Xu \etal \cite{XuICB16}\\(2016)} & \tabincell{l}{Modified AlexNet\\(tattoo vs. non-tattoo)} & \tabincell{l}{n/a} & \tabincell{l}{Tatt-C (tattoo vs. non-tattoo) \\($1,349$; $1000$)$^4$;\\ Flickr (tattoo vs. non-tattoo)\\ ($5,740$; $4,260$)$^4$} & \tabincell{l}{Tatt-C (tattoo vs. non-tattoo): 98.8\%\\ \\ Flickr (tattoo vs. non-tattoo): 78.2\%} \\
\cline{1-5}

\tabincell{c}{Sun \etal \cite{SunICPR16}\\(2016)} & \tabincell{l}{Faster R-CNN} & \tabincell{l}{n/a} & \tabincell{l}{Tatt-C: tattoo vs. non-tattoo \\($1,349$; $1000$)$^4$;\\ Flickr: tattoo vs. non-tattoo\\ ($5,740$; $4,260$)$^4$} & \tabincell{l}{Tatt-C (tattoo vs. non-tattoo): 98.25\% \\ Tatt-C (localization): 45\%@0.1FPPI \\ Flickr (tattoo vs. non-tattoo): 80.66\%} \\
\cline{1-5}

\tabincell{c}{Di and Patel \cite{DiCVPRW16,DiSpringer17}\\(2016)} & \tabincell{l}{AlexNet and SVM \\ (tattoo vs. non-tattoo)} & \tabincell{l}{Siamese network with triplet or \\contrastive loss} & \tabincell{l}{Tatt-C: tattoo vs. non-tattoo \\($1,349$; $1,000$)$^4$;\\ mixed media ($181$; $55$)} & \tabincell{l}{Tattoo vs. non-tattoo: 99.83\%\\
Mixed media: 56.9\% acc.@rank-10} \\
\cline{1-5}

\tabincell{l}{Proposed approach} & \multicolumn{2}{c}{\tabincell{c}{Deep end-to-end learning for joint detection and \\compact representation learning}} & \tabincell{l}{Tatt-C: \\ detection: $7,526$ images \\identification ($157$; $4,375$);\\\\ Flickr (detection): \\$5,740$ images;\\\\ DeMSI (identification): \\$890$ images;\\\\ WebTattoo ($500$, $\sim$$300K$)} & \tabincell{l}{
\textbf{Detection (localization)}\\
Tatt-C: $61.7\%$ recall@0.1FPPI\\
WebTattoo: $87.1\%$ recall@0.1FPPI\\
\textbf{Tattoo search}\\
WebTattoo (photo): \\
$60.1\%$ mAP (w/o background)\\
$25.3\%$ mAP ($300$K background)\\
WebTattoo (sketch): \\
$37.2\%$ mAP (w/o background)\\
\textbf{Tattoo identification}\\
WebTattoo:\\
$63.5\%$ acc.@rank-1 (w/o background) \\
$28.0\%$ acc.@rank-1 ($300$K background) \\
Tatt-C: $99.2\%$ acc.@rank-1 \\
}\\


\bottomrule
\end{tabular}
\end{center}
\vspace{-1mm}
\footnotesize
$^1$Twenty different image transformations were applied to $2,157$ tattoo images to generate $43,140$ synthetic tattoo images.
$^2$Forty different image transformations were applied to $100$ tattoo images to generate $40,000$ synthetic tattoo images.
$^3$$40,000$ images were randomly selected from the ESP game dataset to populate the tattoo dataset.
$^4$(a, b) denotes the number of positive and negative tattoo images per class.
$^5$(a, b) denotes the number of tattoo and non-tattoo images.

\vspace{-4mm}
\end{table*}

These shortcomings of keyword-based tattoo image retrieval systems have motivated the development of content-based image retrieval (CBIR) techniques to improve the tattoo search efficiency and accuracy \cite{Jain07,LeeCVPR08,ActonSSIAI08,JainICIP09,LeeMM12,MangerCRV12,HanICB13}.
CBIR aims to extract features, \eg edge, color, and texture, that can reflect the content of an image, and use them to identify images with high visual similarity.
For example, color histogram and correlogram, shape moments, and edge direction coherence features were used in \cite{Jain07,LeeCVPR08} for tattoo matching.
Similarly, global and local features of edge and color were used in \cite{ActonSSIAI08}, and vector-wise Euclidean distance was computed to measure the similarity between two tattoo images.
The bag-of-words (BoW) model \cite{SivicICCV03} using SIFT \cite{LoweIJCV04} features is probably the most popular one among the early CBIR systems for tattoo search \cite{JainICIP09,LiMMRM09,MangerCRV12,LeeMM12,HanICB13}.
Besides SIFT features, LBP-like features and HoG features were also used in \cite{HeflinBTAS12,WilberWACV14} with SVM and random forest classifiers for tattoo classification.
While these CBIR systems are reported to provide reasonably high accuracies on various benchmarks, they require careful handcrafting of feature descriptor, vocabulary size, and indexing algorithm.

With the success of deep learning in many computer vision tasks \cite{LeCunNATURE15}, the focus of CBIR methods is shifting from handcrafted features and models to deep learning based methods \cite{ZhengTAPMI17}.
In particular, AlexNet \cite{KrizhevskyNIPS12}, winning the ImageNet challenge of $2012$, has been successfully used for tattoo vs. non-tattoo classification in \cite{XuICB16,DiCVPRW16,DiSpringer17}.
Faster R-CNN \cite{RenTPAMI17} was used for both tattoo vs. non-tattoo image classification and tattoo localization in \cite{SunICPR16}.
Among these methods \cite{XuICB16,SunICPR16,DiCVPRW16}, only \cite{DiCVPRW16,DiSpringer17} studied the tattoo identification using a Siamese network with triplet loss.

There are some studies on logo or landmark search, which faces similar challenges to tattoo search, \eg geometric deformation, inverted brightness, etc.
Due to these challenges, image search algorithms based on the traditional BoW representation often fail.
To resolve these issues, great efforts have been made to improve the robustness \cite{ZhouTPAMI18,XieICIP14} and efficiency \cite{ZhouACMMM12,XieICMR15} of the descriptors.
Due to space limitation, we refer interested readers to recent reviews of general image retrieval, \eg \cite{ZhengTAPMI17}.

The current practice of tattoo matching in the literature is towards tattoo identification, and not learning a compact representation for efficient large-scale tattoo search.
Besides, most of the existing tattoo identification methods (including the logo or landmark search algorithms) focus on matching cropped instances (where instance of interest has been segmented from the background), which is different from real application scenarios, where the images are usually uncropped.
Even for the deep learning based methods such as \cite{XuICB16,SunICPR16,DiCVPRW16}, none has addressed tattoo detection and compact representation learning jointly.
The most related work of joint detection and representation learning was reported in face recognition \cite{ChiarXiv17}, in which two large-scale face datasets, \ie WIDER FACE \cite{YangCVPR2016} (containing 393,703 face bounding boxes in 32,203 images, and an average of 12 faces per image) and CASIA WebFace \cite{YiarXiv2014} (containing 494,414 face images of 10,575 subjects), were used to learn their end-to-end face detection and recognition network.
Given the large number of faces per image on average, the small batch size is not an issue in training the recognition part of their joint detection and recognition network.
In contrast, there is usually a single tattoo instance in each image in most of the tattoo datasets; this results in small batch size issue in training the recognition part of our joint tattoo detection and representation learning network because there is only one input image in each iteration. These unique challenges require design of novel end-to-end detection and representation learning approach. In addition, while we aim for efficient large-scale tattoo search and sketch-based tattoo search by performing detection and compact representation learning jointly, \cite{ChiarXiv17} did not report results in such a scenario.

\subsection{Compact Representation Learning}
\label{Sec.CompactRepresentation}

Compact representation learning is of particular interest because of the need for efficient methods in large-scale visual search and instance retrieval applications \cite{NorouziICML11,LiongCVPR15,JainICCV17,ZhengTAPMI17,YangTPAMI17,LuTIP17}.
Compared with high-dimensional real-valued representations, compact representations aim to obtain a compressive yet discriminative feature.
Feature indexing, \ie through various quantization or hashing functions, is a major approach to obtain compact representations.

Quantization based feature indexing methods are designed to quantize the original real-valued representation with minimum quantization errors, and thus usually have high search accuracy \cite{JegouTPAMI11,BabenkoCVPR14,ZhangICML14,WangCVPR16,YuICCV17,ZhouTPAMI18}.
Compared with quantization based methods, hashing based feature indexing methods generates binary codes, providing faster retrieval speed since the Hamming distance of two binary codes can be computed via the native bit-wise operations.
The published hashing based methods for compact representation can be categorized into two major classes: unsupervised and supervised hashing.

Unsupervised hashing algorithms, \eg \cite{IndykSTOC98,DatarSCG04,WeissNIPS08,KulisICCV09,LiuICML11,KongNIPS12,JiangIJCAI15} use unlabeled data to generate binary codes which aim to preserve the similarity information in the original feature space.
Unsupervised hashing methods are often efficient in computation, but their performance in large-scale retrieval may not be the optimum since no label information, including weak labels such as pairwise relationship about a dataset is utilized.
To address this limitation, supervised hashing approaches, \eg \cite{KulisTPAMI09,LiuCVPR12,LinCVPR13,ShenCVPR15,GuiTPAMI17} have been proposed to learn more discriminative binary codes by leveraging both the label similarity and semantic similarity
in the feature values of the data and label information.
Deep neural networks have also been used to learn compact binary codes from high-dimensional inputs \cite{SalakhutdinovICAIS07,KulisNIPS09,LiongCVPR15,LiIJCAI16}.
﻿With the advances in CNN architectures and fine-tuning strategies, the performance of the deep hashing methods is improving, and has provided good generalization ability into new datasets \cite{ZhengTAPMI17}.
Due to the limited space, we refer readers to \cite{WangPIEEE16,ChiCSUR17} for a survey of data hashing approaches. 	

While there are a large number of approaches on hashing based compact representation learning, most of the published methods assume pre-cropped images of instances; but in a fully automatic instance retrieval system, such an assumption usually does not hold.
In addition, most of the published methods on compact representation learning are designed for computer vision tasks such as face image or natural image retrieval, their performance in large-scale tattoo search is not known.

\begin{figure*}[t]
\centering
\includegraphics[width=.95\linewidth]{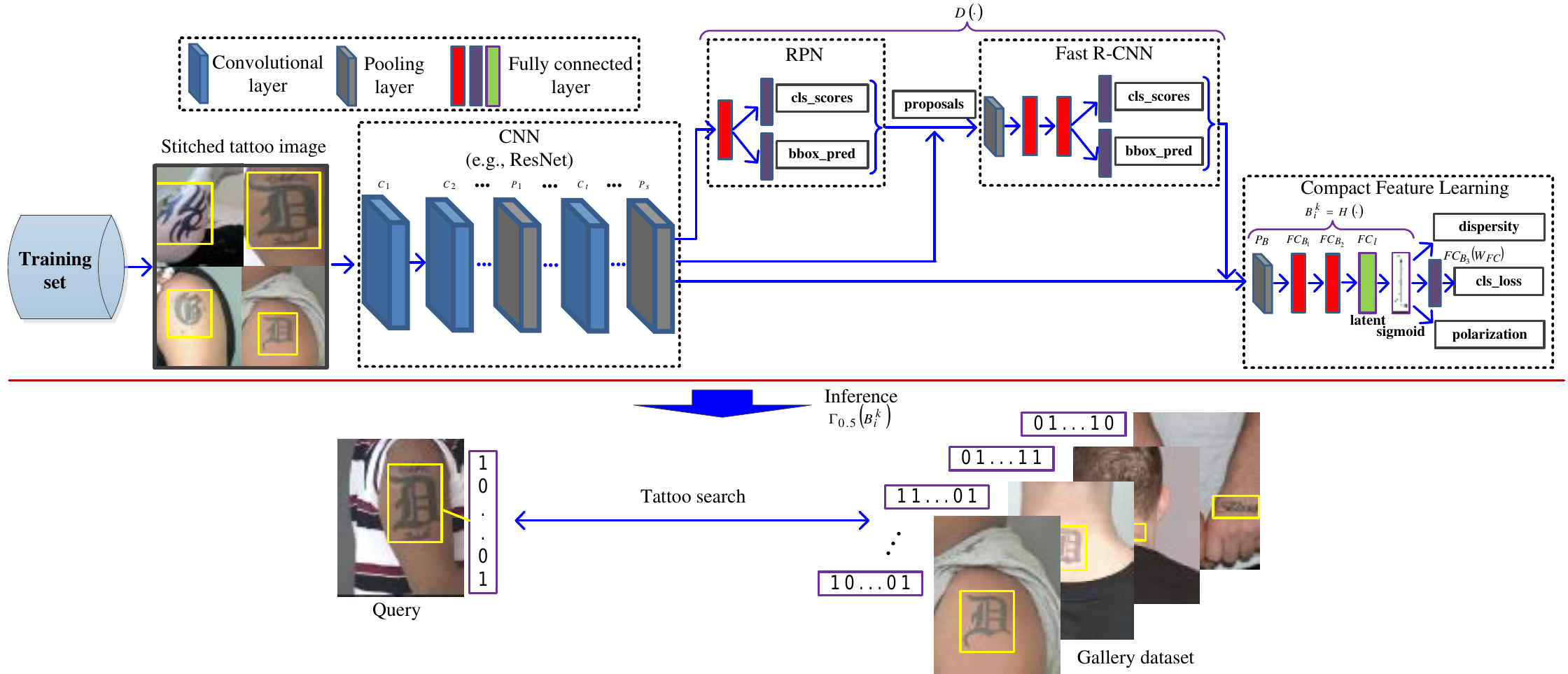}
\vspace{-3mm}
\caption{Overview of the proposed approach for tattoo search at scale via joint tattoo detection and compact representation learning. Our approach consists of a stem CNN for computing the shared features, an RPN \cite{RenTPAMI17} and Fast R-CNN \cite{GirshickICCV15} for tattoo detection, and a compact representation learning module. The proposed approach can be trained end-to-end via stochastic gradient descent (SGD) \cite{KrizhevskyNIPS12} and back-propagation (BP) \cite{RumelhartNATURE86}.}
\label{Fig.JointDetFL}
\vspace{-4mm}
\end{figure*}

\section{Proposed Method}
\label{Sec.ProposedApproach}

\subsection{Review of Faster R-CNN}
\label{Sec.FasterRCNN}

Faster R-CNN \cite{RenTPAMI17} is one of the leading object detection frameworks to identify instances of objects belonging to certain classes and localize their positions (bounding boxes) in an end-to-end learning network.
Faster R-CNN consists of two modules.
The first module, called the Region Proposal Network (RPN), is a fully convolutional network for generating regions of interest (RoI) that denote the possible presence of objects.
The second module is Fast R-CNN \cite{GirshickICCV15}, whose purpose is to classify the RoI by RPN into individual classes and refine the positions of each foreground instance.
By sharing the deep features of the full image between RPN and Fast R-CNN, Faster R-CNN is able to perform object detection accurately and efficiently, and can be trained in an end-to-end fashion.

As aforementioned, conventional methods usually break down the tattoo search problem into two separate tasks, \ie tattoo detection \cite{XuICB16,SunICPR16}, and tattoo matching \cite{LeeCVPR08,HanICB13,WilberWACV14}.
Such a scheme is not optimum because the matching task could assist in the detection task, and the detection accuracy influences the feature discriminability used by the matching task.
Therefore, while Faster R-CNN provides an efficient solution for object detection from images, it addresses only the front-end detection problem of an instance retrieval system.
Our observation is that the convolutional feature maps used by RPN and Fast R-CNN can also be used for learning compact representation.





\subsection{Joint Detection and Compact Representation Learning}
\label{Sec.JointDet&Feature}

We aim to handle tattoo detection and compact representation learning simultaneously via a single model (see Fig. \ref{Fig.JointDetFL}).
A straightforward method for handling tattoo detection and compact representation learning jointly is to use a cascade of tattoo detection and compact representation learning, \ie the output of the detector is fed into the succeeding feature extraction module.
However, such a cascaded method does not leverage feature sharing to achieve efficient and robust representation learning.

Formally, let $\mathbf{A} = \{\mathbf{X}, \mathbf{Y}\}$ be a training tattoo dataset, where $\mathbf{X} = \{\mathbf{X}_i\}_{i=1}^{N}$ denotes $N$ tattoo images from $K$ distinct tattoos, and $\mathbf{Y} = \left\{\{\mathbf{Y}_{i}^{j}\}_{j=1}^{M}\right\}_{i=1}^{N}$ denotes the $M$ labels for the corresponding tattoo images.
Here, the label of each tattoo image consists of two elements (thus $M = 2$), \ie the tattoo position ($\left\{\mathbf{Y}_{i}^{1}\right\}_{i=1}^{N}$) and class ID ($\left\{\mathbf{Y}_{i}^{2} \right\}_{i=1}^{N}$).

Given such a training dataset, we expect to jointly optimize a tattoo detector $\mathbf{D}(\cdot)$ and a compact representation learning function $\mathbf{H}(\cdot)$, which can minimize a regression loss ($\ell_{reg}$) between the predicted and ground-truth bounding boxes, and a classification loss ($\ell_{cls}$) of the compact representations describing individual detected tattoos, respectively.
For the regression loss ($\ell_{reg}$), we choose to use the robust smooth $L_1$ loss \cite{GirshickICCV15}
\begin{equation}
\begin{aligned}
\ell_{reg}\left(\mathbf{D}(\mathbf{X}_{i}), Y_{i}^{1}\right) = \sum_{d \in \{u,v,w,h\}}\mathcal{S}_{L_1}(\mathbf{D}(\mathbf{X_{i}})^{\{d\}} - \mathbf{Y}_{i}^{1,d})
\end{aligned},
\label{Eq:SmoothL1}
\end{equation}
where the four-tuple $\{u,v,w,h\}$ specifies the top-left location $(u, v)$ and the width and height $(w, h)$ of a detected tattoo, and the four elements are indexed by $d$.
The function $\mathcal{S}_{L_1}(\cdot)$ is defined as
\begin{equation}
\begin{aligned}
\mathcal{S}_{L_1}(z) =
\begin{cases}
0.5z^2    & \text{if $|z| < 1$} \\
|z|-0.5 & \text{otherwise}
   \end{cases}
\end{aligned}.
\end{equation}
For the classification loss ($\ell_{cls}$), defined w.r.t. the detected tattoos, we choose to use the cross-entropy loss \cite{deBoerAOR05}
\begin{equation}
\begin{split}
\ell_{cls} = - \sum_{i=1}^{N}\sum_{k=1}^{K}\mathbf{1}\left(\mathbf{\hat{Y}}_{i}^{k}, Y_{i}^{2}\right) \log p\left(\mathbf{\hat{Y}}_{i}^{k}\right)
\end{split},
\label{Eq:CrossEntropy}
\end{equation}
where $\mathbf{\hat{Y}}_{i}^{k} = W_{FC}\cdot\mathbf{B}_{i}^{k} = W_{FC}\cdot\mathbf{H}\left(X_{i}, \mathbf{D}(X_{i})\right)^{\{k\}}$ denoting the $k$-th element of the output by a fully connected layer with weight $W_{FC}$, which takes feature $\mathbf{B}_{i}^{k}$ as its input.
$\mathbf{H}(\cdot)$ takes image $X_{i}$ and the detected tattoo location $\mathbf{D}(X_{i})$ as input, and outputs $\mathbf{B}_{i}^{k}$.
$\mathbf{1}\left(\mathbf{\hat{Y}}_{i}^{k}, Y_{i}^{2}\right)$ outputs $1$ when $k=Y_{i}^{2}$, and $0$ otherwise.
The probability $p(\cdot)$ is computed as
\begin{equation}
\begin{split}
p(\mathbf{\hat{Y}}_{i}^{k}) = \frac{e^{\mathbf{\hat{Y}}_{i}^{k}}}{\sum_{k=1}^{K} e^{\mathbf{\hat{Y}}_{i}^{k}}}
\end{split}.
\label{Eq:Softmax}
\end{equation}

By minimizing the losses given in (\ref{Eq:SmoothL1}) and (\ref{Eq:CrossEntropy}), we can jointly perform tattoo detection and feature representation learning from the detected tattoos.
However, additional constraints are still required to guarantee that the learned features are compact binary features, which is important for efficient large-scale search.
Therefore, we expect that the features, \ie $\mathbf{B}_{i}^{k} = \{b_k|b_k \in \{0, 1\}\}_{k=1}^{K}, i=1,2,\cdot\cdot\cdot, N$, learned by $\mathbf{H}(\cdot)$ should be near-binary codes.
This implies that each element $\mathbf{B}_{i}^{k}$ of a feature vector should be close to either $1$ or $0$.
Such an objective can be approximated by penalizing the learned feature to have elements close to $0.5$
\begin{equation}
\begin{split}
\ell_{pol}(\mathbf{B}_{i}) = \frac{1}{\frac{1}{2K}\sum_{k=1}^{K}\|\mathbf{B}_{i}^{k} - 0.5\|_2^2 + \epsilon},
\end{split}
\label{Eq:Polarization}
\end{equation}
in which $\epsilon$ is a small positive constant for avoiding divide-by-zero, and we use $\epsilon = 0.01$ in our experiments.
We call such a loss in (\ref{Eq:Polarization}) as a \emph{polarization} loss.

\begin{figure}[t]
\centering
\includegraphics[width=0.8\linewidth,height=0.6\linewidth]{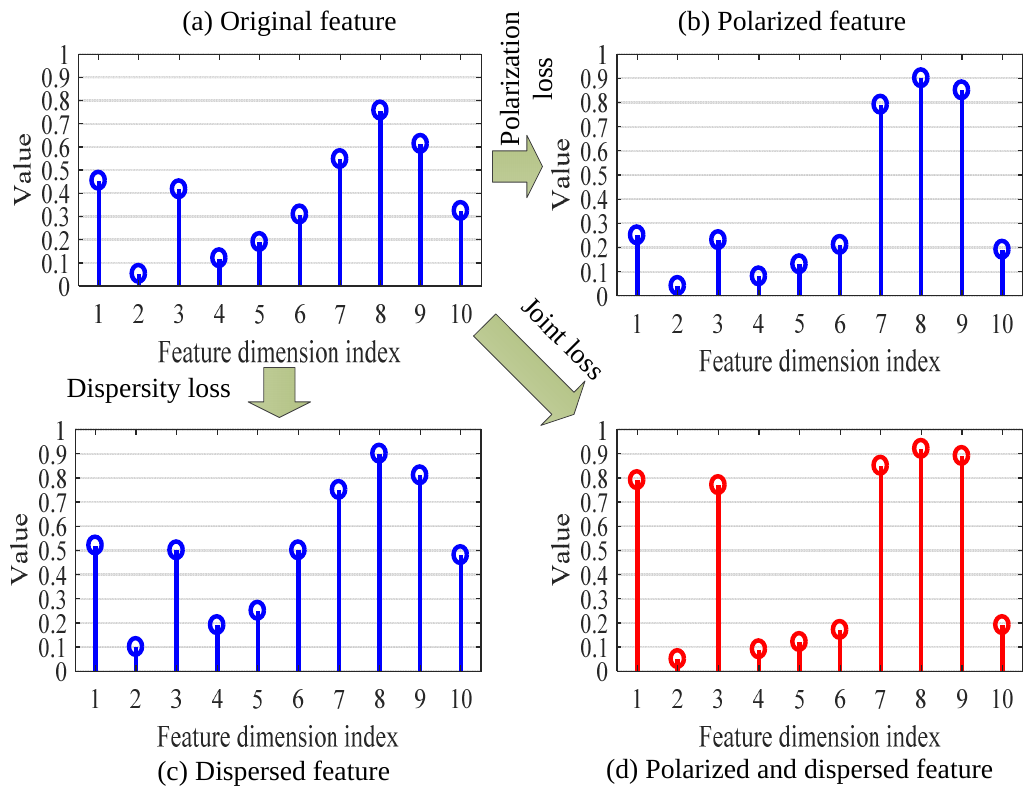}
\vspace{-4mm}
\caption[Polarization]{The benefit of using the polarization loss and dispersity loss. (a) an original feature vector of real values in the range of $[0, 1]$, (b) the learned feature vector after using the polarization loss alone (\ie each element is close to either $0$ or $1$), (c) the learned feature vector after using the dispersity loss alone (\ie the elements are evenly distributed on the two sides of $0.5$), and (d) the learned feature vector after jointly using the polarization and dispersity losses (\ie near-binary elements are evenly distributed on the two sides of $0.5$).}
\label{Fig.PolarizationLoss}
\vspace{-4mm}
\end{figure}

In addition, we expect that every bit in a binary code of length $\iota$ can contribute to the representation of individual tattoos.
In other words, each bit of the binary code is expected to have a $50\% $ fire rate \cite{YangTPAMI17}.
Such an objective can be approximated by constraining the average value of a learned feature to be $0.5$, \ie
\begin{equation}
\begin{split}
\ell_{dis}(\mathbf{B}_{i}) = \frac{1}{\iota}\sum_{k=1}^{\iota}\mathbf{B}_{i}^{k} - 0.5
\end{split}
\label{Eq:Dispersity}
\end{equation}
We call such a loss in (\ref{Eq:Dispersity}) as a \emph{dispersity} loss.

By minimizing the losses defined in (\ref{Eq:CrossEntropy}), (\ref{Eq:Polarization}) and (\ref{Eq:Dispersity}) jointly, the learned features are expected to be discriminative near-binary codes that are evenly distributed in the feature space (see Fig. \ref{Fig.PolarizationLoss}).
Such near-binary codes can be easily converted into a binary code utilizing a threshold function
\begin{equation}
\begin{aligned}
\mathbf{B'}_{i}^{k} = \mathcal{T}(\mathbf{B}_{i}^{k}) =
\begin{cases}
0    & \text{if $\mathbf{B}_{i}^{k} < 0.5$} \\
1    & \text{otherwise}
\end{cases}
\end{aligned}.
\label{Eq:Binaryzation}
\end{equation}
Finally, we compute the distance between a query tattoo image and a gallery tattoo image using the Hamming distance of the binary vectors.
In case more than one tattoos are detected from an image, \ie $u$ and $v$ tattoos are detected from a query image and a gallery image, respectively, we compute $u \times v$ distances in total, and use the minimum distance as the final distance between the two tattoo images.

\subsection{Network Structure}

While our aim of joint tattoo detection and compact representation learning is well defined by (\ref{Eq:SmoothL1}), (\ref{Eq:CrossEntropy}), (\ref{Eq:Polarization}), and (\ref{Eq:Dispersity}), the feature sharing between detection and representation learning tasks is not simple.

We propose to perform joint tattoo detection and compact representation learning based on a Faster R-CNN model by embedding the above four losses into a single network.
Specifically, a CNN such as AlexNet \cite{KrizhevskyNIPS12}, VGG \cite{SimonyanArXiv2014}, or ResNet \cite{HeCVPR16} can be used to extract the deep features that are to be shared by RPN and Fast R-CNN.
The RPN and Fast R-CNN modules, and the tattoo regression loss defined in (\ref{Eq:SmoothL1}) in the proposed approach are the same as those in the original Faster R-CNN (see Fig. \ref{Fig.JointDetFL}).

We establish joint compact representation learning by introducing a new compact representation learning (CRL) sub-network (see Fig. \ref{Fig.JointDetFL}).
CRL consists of an instance pooling layer (\eg $P_B$), a sequence of fully connected (FC) layers (\eg $FC_{B_1}$ and $FC_{B_2}$), a latent layer with sigmoid activation (\eg $FC_{l}$), and three sibling output layers (\eg cls\_loss, polarization, and dispersity).
The instance pooling layer and the sequence of FC layers are the same as the RoI pooling and FC layers in Fast R-CNN, which compute a fixed-length feature vector from the shared feature map for each tattoo detection by Fast R-CNN, and optimize this feature w.r.t. the following CRL tasks.
The latent layer (also an FC layer) with sigmoid activation is expected to generate near-binary representation $\mathbf{B}_{i}$ that will be binarized via (\ref{Eq:Binaryzation}) for large-scale search.
The three sibling output layers model the corresponding constraints, \ie polarization loss in (\ref{Eq:Polarization}), dispersivity loss in (\ref{Eq:Dispersity}), and classification loss in (\ref{Eq:CrossEntropy}), respectively (see Fig. \ref{Fig.JointDetFL}).
Overall, the three loss functions work in a multi-task way to perform CRL given a tattoo detection.
We use hyper-parameters to control the balance between individual losses
\begin{equation}
\begin{split}
\ell_{J}(\mathbf{B}_{i}) = \alpha\ell_{cls} + \beta\ell_{pol} + \gamma \ell_{dis}
\end{split}
\label{Eq:JointLoss}
\end{equation}
We set $\alpha=\beta=\gamma=1$ based on empirical results.
It should be noted that although there are $K$ classes of tattoos in the training dataset, the number of outputs in both RPN and Fast R-CNN remains two (corresponding to background and tattoo).
The number of outputs for classification loss layer in our CRL is $K$.

\begin{figure*}[t]
\centering
\includegraphics[width=.9\linewidth]{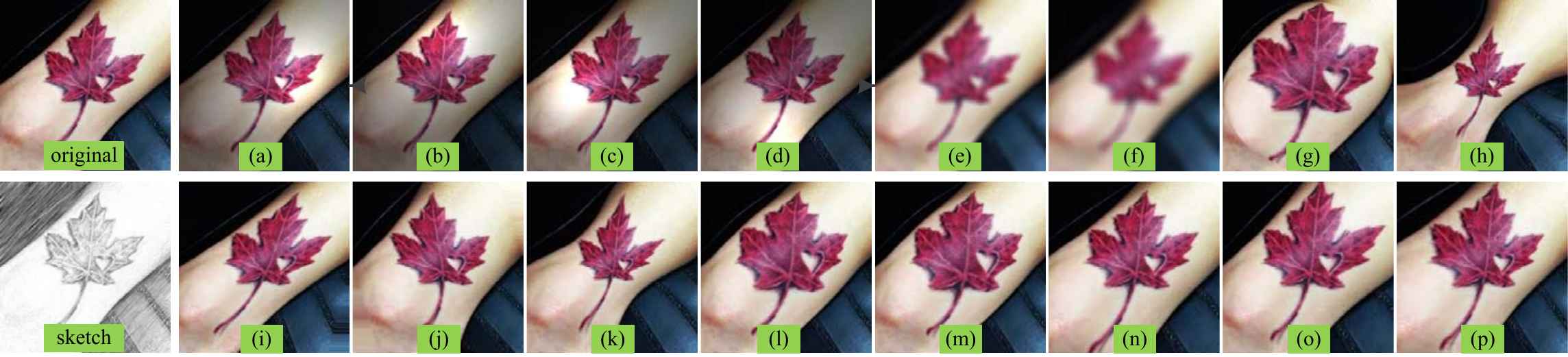}
\vspace{-3mm}
\caption{An example of data augmentation for one tattoo image (referred to as ``original'' in the top row) in the training set to replicate various image acquisition conditions, \eg (a-d) illumination variation, (e-f) image blur, (g-l) deformation, and (m-p) perspective distortion. A tattoo sketch is also generated for data augmentation (shown under the original tattoo).}
\label{Fig.TattooAugmentation}
\vspace{-4mm}
\end{figure*}

The proposed approach differs from \cite{SalvadorCVPRW16,XiaoCVPR17} in that: (i) \cite{SalvadorCVPRW16} optimizes the feature representation for instance search by fine-tuning the detection network w.r.t. the query instances; however such features optimized for detection tasks may not be optimal for retrieval tasks.
By contrast, the proposed approach jointly optimizes both detection and feature representation end-to-end;
(ii) while the features used in \cite{SalvadorCVPRW16,XiaoCVPR17} are real-valued, the proposed approach learns compact binary features, which are scalable; (iii) the bounding boxes used for computing the features for instance matching in \cite{XiaoCVPR17} are not accurate compared to the final bounding box estimates used by the proposed approach; however, as we will show in the experiments, accurate tattoo bounding boxes are important for improving the tattoo matching accuracy; (iv) while \cite{XiaoCVPR17} studied person search with a gallery set containing about $6,978$ images, the scalability of the proposed approach is studied with a gallery dataset containing more than $300$K distracter tattoo images; and (v) while cross-modality instance search, \ie sketch based tattoo search is studied in our work, the performance of \cite{XiaoCVPR17} under a cross-modality matching scenario is not known.


\subsection{Implementation Details}
\label{Sec.ImpDetails}

\textbf{Data Augmentation.} Compared with the large-scale databases for object detection \cite{LinarXiv14} and image classification \cite{RussakovskyIJCV15}, the public-domain tattoo datasets such as Tatt-C \cite{NganNIST15}, Flickr \cite{XuICB16}, and DeMSI \cite{HrkacVMR16} are of limited sizes (usually less than $10K$).
The limited datase size poses additional challenges to the proposed approach, \ie the risk of overfitting, particularly for our CRL module, which usually requires multiple tattoo images per class and a large number of tattoo classes to learn a robust model.
Besides the commonly used data augmentation methods (\eg random crop, translation, rotation, and reflection) \cite{KrizhevskyNIPS12}, we have designed $16$ additional transformations\footnotemark[7] to replicate the diversity of one tattoo instance caused by various acquisition conditions (see Fig. \ref{Fig.TattooAugmentation}).
In addition, we have also generated a tattoo sketch (see Fig. \ref{Fig.TattooAugmentation}) for data augmentation so that the proposed approach can generalize to sketch-based tattoo retrieval task.
\footnotetext[7]{All the augmentations were performed leveraging the Photoshop plug-ins for MATLAB: \url{https://helpx.adobe.com/photoshop/kb/downloadable-plugins-and-content.html}}


\textbf{Network Training.} We use ResNet-50 as our backbone network for shared feature learning, which is pretrained on ImageNet \cite{RussakovskyIJCV15} for parameters initialization.
We use a confidence threshold of $0.8$ to filter the tattoo detections by Fast R-CNN, which corresponds to about one false detection for every ten images, on average.
The filtered tattoo detections are fed into CRL.
The reason why we use a relatively high threshold is to avoid feeding non-tattoo detections into CRL, wich may cause difficulty in network convergence.
We use a learning rate of $10^{-4}$ during the fine-tuning of the pre-trained CRL.
For the parameters of RPN and Fast R-CNN, we directly use the suggested values in \cite{GirshickICCV15,RenTPAMI17}.

Since our approach performs tattoo detection and CRL jointly, the detection module usually takes one image as input (given a typical Titan X GPU), and outputs one detected tattoo, which is then used as the input to CRL.
Thus, the batch size w.r.t. CRL is limited to one tattoo, which makes it difficult for CRL to converge.
Such an issue cannot be resolved by just compiling a large training dataset or using data augmentation as in Fig. \ref{Fig.TattooAugmentation}.
To address this issue, we make use of preceding feature buffering \cite{XiaoCVPR17} to assist in CRL training.
In addition, we stitch multiple randomly selected training images into a single image (see Fig. \ref{Fig.JointDetFL}), and use it as the input to our detection module so that it can output multiple detected tattoos.
In this way, multiple detected tattoos can be used for training CRL, and thereby improving the training batch size.
We have found such stitched tattoo images to be very for training our joint tattoo detection and CRL network.
We also use online hard example mining (OHEM) \cite{ShrivastavaarXiv16} in Fast R-CNN to improve its robustness in detection blurred, partial, and tiny tattoos.
All the tattoo images are scaled before they are input to the network so that the shorter edge between width and height is $600$ pixels.

\section{Experiments}
\label{Sec.Experiment}

\subsection{Databases}
\label{Sec.Databases}

There are only a limited number of tattoo databases in the public domain, such as Tatt-C \cite{NganNIST15}, Flickr \cite{XuICB16}, and DeMSI \cite{HrkacVMR16}.
These tattoo databases are used in the evaluations of our approach and comparisons with state-of-the-art.

\begin{figure*}[t]
\centering
\includegraphics[width=.9\linewidth]{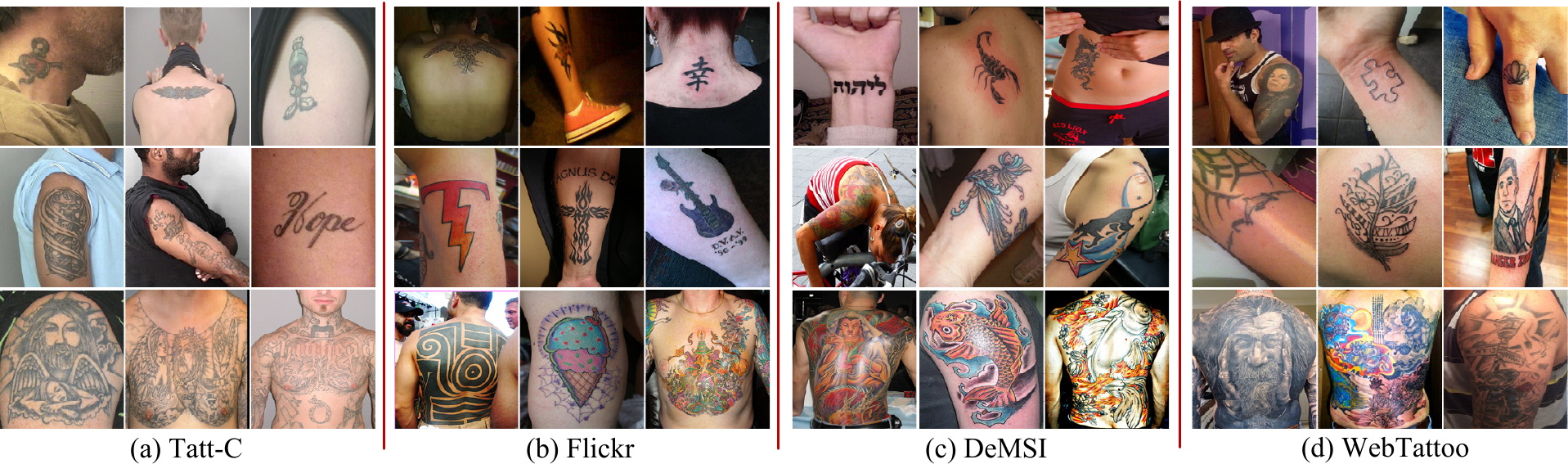}
\vspace{-3mm}
\caption[]{Examples of tattoo images from the four tattoo image databases used in our experiments: (a) Tatt-C \cite{NganNIST15}, (b) Flickr \cite{XuICB16}, (c) DeMSI \cite{HrkacVMR16}, and (d) our WebTattoo dataset.}
\label{Fig.DatabaseExample}
\vspace{-4mm}
\end{figure*}

\emph{Tatt-C.} The NIST Tatt-C database was developed as an initial tattoo research corpus that addresses use cases representative of operational scenarios \cite{NganNIST15}.
The tattoo vs. non-tattoo classification dataset in Tatt-C contains $1,349$ and $1,000$ tattoo and non-tattoo images, respectively.
The tattoo identification dataset in Tatt-C contains $157$ and $215$ probe and gallery images, respectively.
A background dataset with $4,332$ non-tattoo images was also used to populate the gallery set.
The tattoo mixed-media dataset in Tatt-C, consisting of photos, sketches, and graphics, contains $181$ and $272$ probe and gallery images, respectively.
A five-fold cross-validation was used for the identification experiment of each dataset in Tatt-C \cite{NganNIST15}.
We also notice that the bounding-box annotations were provided for $7,526$ tattoo images in Tatt-C, so we also report the tattoo detection accuracy using these image in Tatt-C.
The tattoo images in the Tatt-C dataset contain variations of illumination, partial occlusion, and image blur (see Fig. \ref{Fig.DatabaseExample} (a)).

\emph{Flickr.} The Flickr tattoo database contains $5,740$ and $4,260$ tattoo and non-tattoo images that were collected from Flickr \cite{XuICB16}.
We can notice that the ratio of the tattoo images to the non-tattoo images is similar to that of the Tatt-C database.
While the tattoo images in the Tatt-C database were collected from an indoor environment, the images in the Flickr database were taken from both indoor and outdoor environment, with diverse viewpoints, poses, and complex backgrounds (see Fig. \ref{Fig.DatabaseExample} (b)).
The Flickr database was original built for tattoo vs. non-tattoo classification.
We have extended the Flickr database by providing bounding-box annotations for each tattoo in the images.\footnotemark[8]
Therefore, in our experiments, we are able to use the Flickr database to evaluate the tattoo detection (localization) performance.
\footnotetext[8]{We will put the bounding-box annotations for the Flickr database into the public domain.}

\emph{DeMSI.} The DeMSI dataset contains $890$ tattoo images from the ImageNet \cite{RussakovskyIJCV15} database.
The boundary of each tattoo image was annotated for tackling the tattoo segmentation problem \cite{HrkacVMR16}.
Since the tattoo images are from the ImageNet database, they are captured under an unconstrained scenario (see Fig. \ref{Fig.DatabaseExample} (c)).
The tattoo images from the DeMSI dataset are used together with our WebTattoo databases as the background images to populate the gallery dataset.

\emph{WebTattoo.} We can notice that the above tattoo datasets are usually of limited sizes (less than $10K$).
Although the NIST Tatt-E challenge is reported to have a much larger tattoo testing dataset collected from real application scenarios\footnotemark[9], there is no evidence this dataset will be put into the public domain.
\footnotetext[9]{\url{https://www.nist.gov/programs-projects/tattoo-recognition-technology-evaluation-tatt-e}}
To replicate the operational scenario of tattoo search at scale, we have compiled a large tattoo database (named as WebTattoo) by (i) combining the above three public-domain tattoo databases together, (ii) collecting over than $300$K distracter tattoo images from the Internet (see Fig. \ref{Fig.DatabaseExample} (d)), and (iii) drawing $300$ tattoo sketches by volunteers, who were asked to take a look at a tattoo image for one minute and then draw the tattoo sketch the next day (see an example of the tattoo sketch in Fig. \ref{Fig.TattooAugmentation}).\footnotemark[10]
\footnotetext[10]{We plan to put the WebTattoo dataset into the public-domain.}

\begin{figure*}[t]
\centering
\includegraphics[width=.90\linewidth]{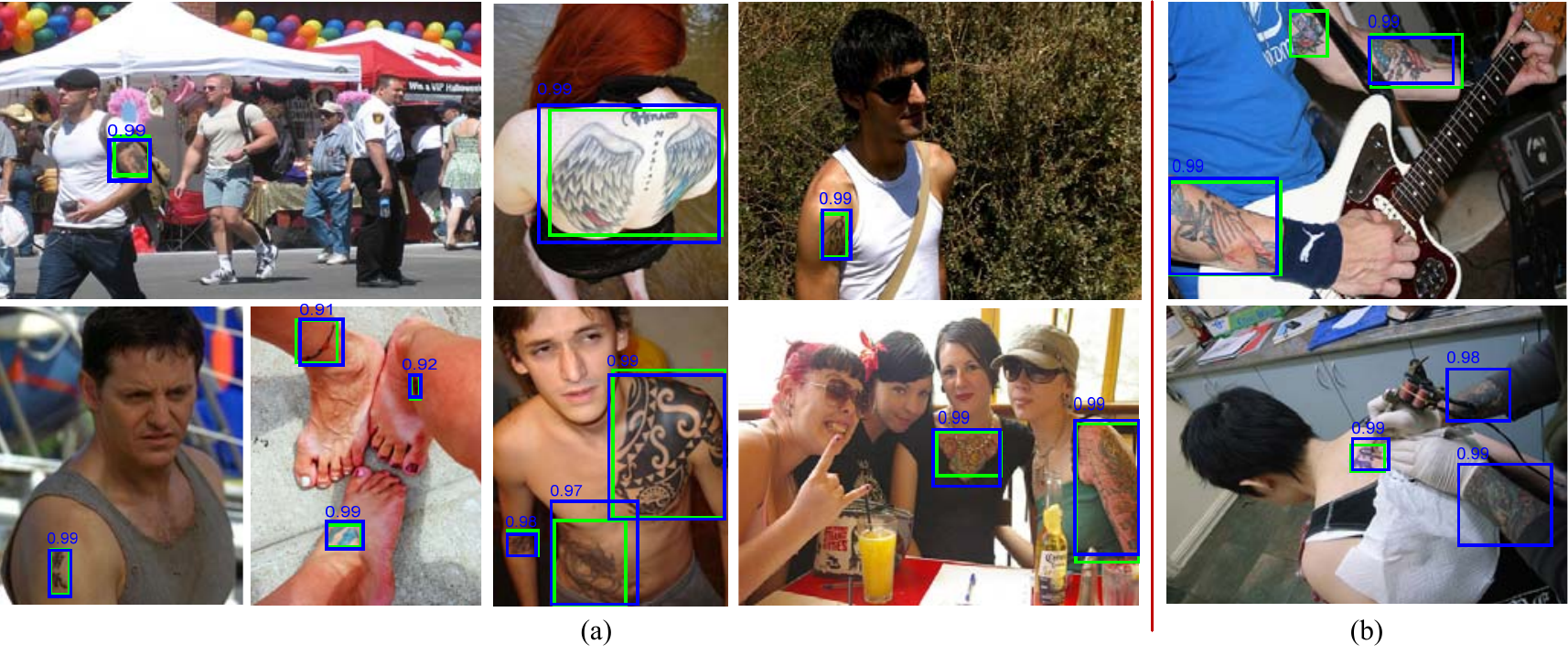}
\vspace{-4mm}
\caption[]{Examples of (a) good tattoo detections, and (b) poor tattoo detections by the proposed approach. The green and blue rectangles show the ground-truth tattoo bounding boxes and the detected tattoo bounding boxes, respectively. The numbers shown above the bounding boxes are the detection confidence scores.}
\label{Fig.DetectionExamples}
\vspace{-4mm}
\end{figure*}

Based on the WebTattoo dataset, we use a semi-automatic approach to find the tattoo classes which have multiple tattoo images per class.
Specifically, a ResNet-50 network pre-trained on ImageNet is first used for automatic tattoo feature extraction and clustering (we used k-means clustering \cite{JainCSUR99}).
The clusters are then manually verified to assure that each cluster contains only one class of tattoo images.
Finally, we obtained about $600$ tattoo classes, with nearly three tattoo images per class on average.
We randomly choose about $1,400$ tattoo images from $400$ tattoo classes for training, and use tattoo images of the remaining $200$ tattoo classes for testing.
The training set is augmented using the method described in Sect. \ref{Sec.ImpDetails}.
We have manually annotated the tattoo bounding boxes for more than $78$K tattoo images (original and augmented tattoo images) in total.
Given such a large number of annotations, there are inevitably some missed tattoos by human workers (see Fig. \ref{Fig.DetectionExamples} (b)).
However, such issues are not unique in our tattoo search task; they exist in many databases for individual computer vision tasks, such as face detection and recognition, person detection and recognition, etc.
In addition to the data augmentation, we also use the $5,740$ tattoo images from the Flickr dataset for training.
Since no class label is provided for the tattoo images in Flickr, these tattoo images contribute only to the detection loss in the proposed approach during network training.
For each class of tattoos in the testing set, we randomly choose one tattoo image for the query, and use the remaining tattoo images for the gallery.
Overall, we have $200$ tattoo images in the query and $350$ tattoo images in the gallery.
About $300$K WebTattoo images that are not present in the training, gallery, and query sets, are used as the distracter tattoo images to populate the gallery set, and replicate the large-scale tattoo search scenario.
For the $300$ pairs of tattoo sketches and the mated tattoo photos, we randomly choose $240$ pairs for training, and the remaining $60$ pairs for testing.
For each pair of tattoo sketch and image, the tattoo sketch is used for query, and the tattoo image is used for gallery.

An operational tattoo dataset reported in \cite{MangerCRV12} contains $327,049$ tattoo images collected by the German police.
Another operational tattoo dataset reported in \cite{LeeMM12} contains about $64,000$ tattoo images, provided by the Michigan State Police.
Our extended gallery set contains more than $300$K tattoo images, which should reasonably replicate the operational tattoo search scenario.


\subsection{Evaluation Metrics}
\label{Sec.Metrics}

The evaluations of the proposed approach and the comparisons with the state-of-the-art tattoo retrieval and identification methods cover the tasks of tattoo detection, identification, and large-scale search.
For each task, we choose to use the widely used evaluation metric in the literature.

\emph{Tattoo detection.} We use the detection error trade-off (DET) curve to measure the tattoo detection performance, \ie the recall vs. false positives per image (FPPI).
Given an intersection-over-union (IoU) threshold (we use $0.5$) between the detected tattoo bounding boxes and the ground-truth tattoo bounding boxes, recall is defined as the fraction of detected bounding boxes with an IoU to the ground-truth larger than the threshold over the total amount of ground-truth bounding boxes.

\emph{Tattoo search.} We use the precision-recall curve to measure the tattoo search performance.
Precision is the fraction of the mated tattoo images that have been retrieved over all the retrieved results for a given query tattoo.
Recall, similar to that in the detection task, is the fraction of the mated tattoo images that have been retrieved over the total amount of mated tattoo images for a given query tattoo.

\emph{Tattoo identification.} We use the cumulative match characteristic (CMC) curve to measure the tattoo identification performance.
Each point on CMC gives the fraction of the probe tattoo images that are correctly matched to their mated gallery images at a given rank.

\subsection{Tattoo Detection}
\label{Sec.TattooDetection}

Since the proposed approach can perform tattoo detection and compact representation learning jointly, we first evaluate the tattoo detection performance of the proposed approach on the WebTattoo test and Tatt-C datasets.
Specifically, we train our approach using the WebTattoo training set, and report the tattoo detection accuracy on the WebTattoo test and Tatt-C datasets.
Since the Tatt-C dataset was primarily built for tattoo vs. non-tattoo classification and tattoo identification tasks, only a limited number of published methods have reported tattoo detection performance on Tatt-C \cite{SunICPR16}.
To provide more baseline performance, we train a Faster R-CNN tattoo detector used in \cite{SunICPR16} on the WebTattoo training dataset, and report its performance on the WebTattoo test and Tatt-C datasets.
We should note that such a cross-database testing protocol is more challenging than the intra-database testing protocol used in \cite{SunICPR16}.

Table \ref{tab:TattooDetRes} lists the tattoo detection performance of the proposed approach and the baseline methods.
The state-of-the-art tattoo detection method in \cite{SunICPR16} reported about $45\%$ recall @ $0.1$FPPI on a Tatt-C dataset with about $2,000$ tattoo images (one tattoo per image).
The Faster R-CNN tattoo detector we trained gives $56.2\%$ and $80.2\%$ recalls at $0.1$FPPI on the Tatt-C and WebTattoo test datasets, respectively.
The performance of the Faster R-CNN tattoo detector we trained is much higher than that in \cite{SunICPR16}, even though we are using a challenging cross-database testing protocol, and the Tatt-C subset we used for evaluation contains much more tattoo images than that was used in \cite{SunICPR16} ($7,526$ vs. $2,000$ tattoo images).
The possible reason is that the WebTattoo training set contains more tattoo images than those used in the intra-database testing, which is helpful for training a deep learning based tattoo detector.
In addition, our data augmentation can replicate the appearance variations existing in various tattoo images, and thus is helpful to improve the robustness of the tattoo detector in unseen scenarios.
The proposed approach for joint tattoo detection and CRL achieves $61.7\%$ and $87.1\%$ recalls at $0.1$FPPI on the Tatt-C and WebTattoo test datasets, respectively, which are much better than the state-of-the-art tattoo detectors.
The results indicate that the proposed approach can leverage multi-task learning to achieve robust feature learning and detector modeling.
Another baseline tattoo detection method in \cite{KimHST16} used a Graphcut based method, and reported $70.5\%$ precision @ $41.0\%$ recall on a Tatt-C dataset with $6,308$ tattoo images.
Under the same evaluation metric, the proposed approach can achieve $99.0\%$ precision @ $41.0\%$ recall on the above Tatt-C dataset with $7,526$ tattoo images.

\begin{table}[!t]
\begin{center}
\caption{Tattoo detection (localization) performance of the proposed approach and the state-of-the-art methods on the WebTattoo test and Tatt-C datasets in terms of recall vs. FPPI.}
\label{tab:TattooDetRes}
\renewcommand{\arraystretch}{1.1}
\footnotesize
\begin{threeparttable}
\begin{tabular}{>{\centering\arraybackslash}p{1.5cm}>{\centering\arraybackslash}p{1.7cm}>{\centering\arraybackslash}p{1.7cm}>{\centering\arraybackslash}p{1.7cm}}
\toprule
\multirow{2}{*}{\textbf{Method}} & \multicolumn{3}{c}{\textbf{Recalls (in $\%$) @ different FPPIs}} \\
\cline{2-4}
                                 & \tabincell{c}{\textbf{0.01 FPPI}\\\textbf{Tatt-C/WebTatt}} & \tabincell{c}{\textbf{0.1 FPPI}\\\textbf{Tatt-C/WebTatt}} &
                                 \tabincell{c}{\textbf{1.0 FPPI}\\\textbf{Tatt-C/WebTatt}} \\
\midrule
\tabincell{c}{Sun \etal \cite{SunICPR16}$^1$} & $8$/$-$ & $45$/$-$ & $-$/$-$\\
\tabincell{c}{Faster \\R-CNN \cite{RenTPAMI17}$^2$} & $17.1$/$21.7$ & $56.2$/$80.2$ & $72.2$/$94.6$\\
Proposed$^2$     & $\textbf{45.9}$/$\textbf{27.5}$ & $\textbf{61.7}$/$\textbf{87.1}$ & $\textbf{80.0}$/$\textbf{95.5}$ \\
\bottomrule
\end{tabular}
$^1$The results are from \cite{SunICPR16}, in which a Tatt-C dataset with about $2,217$ tattoo images was used. $^2$Similar to \cite{SunICPR16}, we trained a Faster R-CNN tattoo detector using the WebTattoo training set, and tested it on the Tatt-C (with $7,526$ tattoo images) and WebTattoo test datasets. This is a cross-database testing scenario, which is more challenging than that used in \cite{SunICPR16}. We use an intersection-over-union (IoU) threshold of $0.5$ between the detected and ground-truth bounding boxes.
\end{threeparttable}
\end{center}
\vspace{-6mm}
\end{table}





Fig. \ref{Fig.DetectionExamples} shows examples of good and poor tattoo detections by our approach on the WebTattoo database.
We find that the proposed approach is quite robust to large pose and illumination variations as well as the diversity of tattoo categories.
Some of the false detections by the proposed approach are due to the missed labeling of the tattoos (see the bottom tattoo image in Fig. \ref{Fig.DetectionExamples} (b)).
However, we notice that detecting tiny tattoos that are easily confused with the background region remains a challenging problem.

\begin{figure}[t]
\centering
\includegraphics[width=0.78\linewidth]{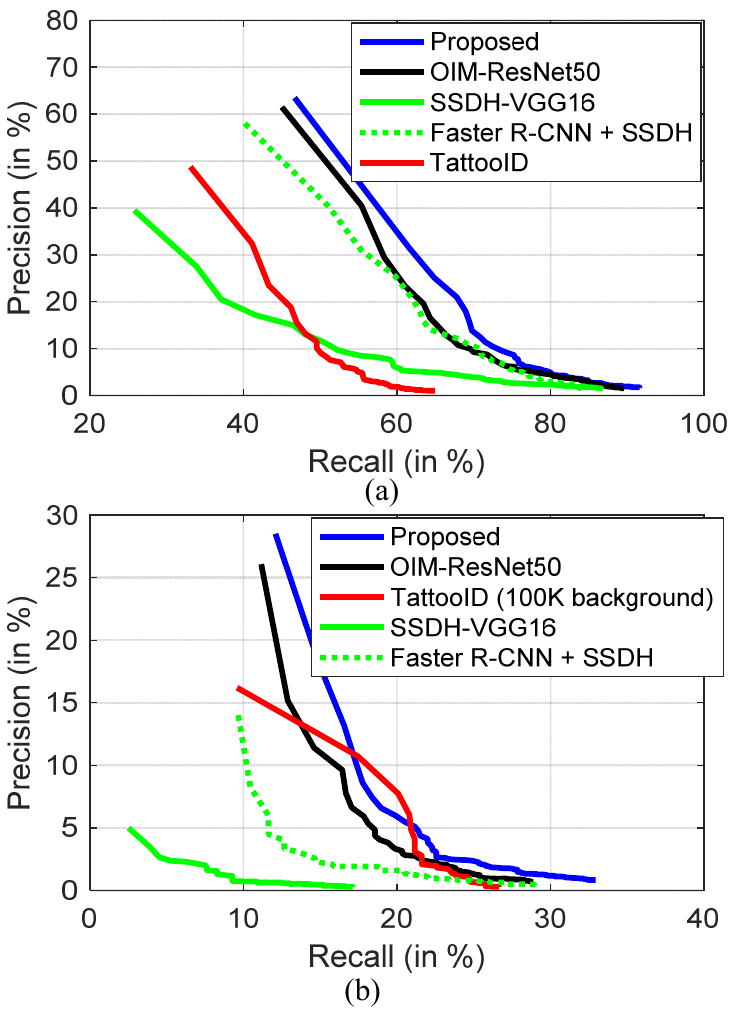}
\vspace{-4mm}
\caption[TattooSearchPR]{Tattoo search performance (in terms of precision-recall) by the proposed approach and the state-of-the-art methods (TattooID \cite{LeeMM12}, SSDH-VGG16 \cite{YangTPAMI17}, Faster R-CNN + SSDH, and OIM-ResNet50 \cite{XiaoCVPR17}) on the WebTattoo test dataset: (a) without background tattoo images in the gallery set, and (b) with $300$K background tattoo images in the gallery set; for TattooID, we report its tattoo search performance using an extended gallery set with only $100$K background images because of its long running time.}
\label{Fig.TattooSearchPR}
\vspace{-4mm}
\end{figure}

\subsection{Tattoo Search}
\label{Sec.SearchatScale}

Efficient tattoo search is important for scenarios, where the search must be operated in a large volume of raw images or video frames.
We evaluate the our approach for tattoo search at scale, and provide comparisons with several state-of-the-art methods \cite{Jain07,LeeMM12,YangTPAMI17,XiaoCVPR17}.
For TattooID \cite{Jain07,LeeMM12}, we reimplement it in Matlab because the early algorithm has been licensed to MorphoTrak\footnotemark[11].
For SSDH-VGG16 \cite{YangTPAMI17} and OIM-ResNet50 \cite{XiaoCVPR17}, we directly use the code provided with their papers, and train the models using the same training set as our approach.
Since SSDH-VGG16 is not able to detect tattoos from an input image, we also consider another baseline, \ie Faster R-CNN is applied for tattoo detection first, and then SSDH-VGG16 is used to extract compact features for the detected tattoos (Faster R-CNN + SSDH).
We have tried several confidence thresholds (\eg $0.3$, $0.5$, $0.7$ and $0.9$) for tattoo detection using Faster R-CNN, and finally chosen to use a threshold of $0.3$, because of its good performance for the final tattoo search.
For both SSDH-VGG16 and OIM-ResNet50, we use $256$D feature representations as suggested in their papers.
For our approach, we also use a $256$-bit compact feature for fair comparisons.
When the gallery size is too large, \ie with $300$K background tattoo images in the gallery, for efficiency, we compute the precision-recall using the top-100 retrieval results.
\footnotetext[11]{https://msutoday.msu.edu/news/2010/msu-licenses-tattoo-matching-technology-to-id-criminals-victims}

\begin{figure}[t]
\centering
\includegraphics[width=0.78\linewidth]{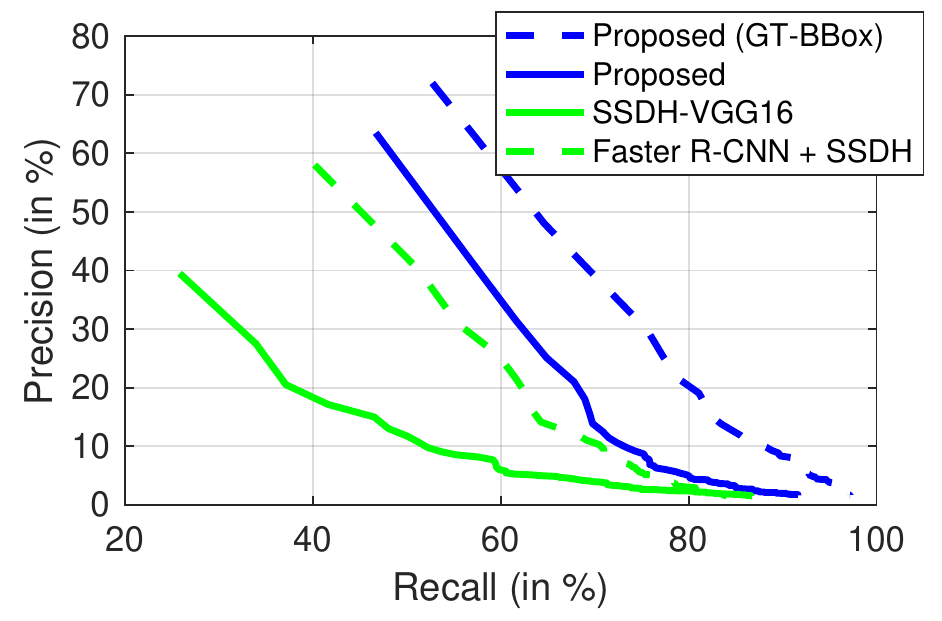}
\vspace{-4mm}
\caption[ImportanceAccurateBBox]{The importance of using accurate tattoo bounding boxes for compact feature learning.}
\label{Fig.ImportanceAccurateBBox}
\vspace{-4mm}
\end{figure}

Fig. \ref{Fig.TattooSearchPR} (a) shows the precision-recall curves of the proposed approach and the state-of-the-art methods for tattoo search without using the $300$K background tattoo images to populate the gallery set.
We are surprised to see that TattooID, a non-learning based matcher based on SIFT features, performs better than the deep learning based method SSDH-VGG16.
The main reason is that SSDH-VGG16 alone is a holistic approach, which learns features from the entire tattoo images.
Since many tattoo images contain large background regions around the tattoos, such a feature representation may capture more characteristics about the background regions than the tattoos, and thus leads to incorrect matches of tattoo images.
This is also the reason why Faster R-CNN + SSDH achieves better performance than SSDH-VGG16.
OIM-ResNet50, which is also a joint detection and feature learning method, is able to leverage the tattoo detection to reduce the influence of the background regions of the tattoos.
As a result, the learned features could better represent the content of a tattoo, and achieves much higher tattoo search accuracy.
However, OIM-ResNet50 extracts the features based on the region proposals of the tattoos instead of the final location estimations of the tattoos.
Such a feature representation is less accurate than the proposed approach, which utilizes the final location estimations of the tattoos.
The proposed joint tattoo detection and CRL approach performs better than all the baseline methods.
The results suggest that multi-task learning used in our approach is helpful for learning more informative representation for tattoo search.
In addition, the proposed approach leverages OHEM to improve the tattoo detection robustness, and stitched training images to increase the instance-level batch size during CRL.
In Fig. \ref{Fig.ImportanceAccurateBBox}, we also provide the performance of tattoo search using the ground-truth tattoo bounding boxes to extract our compact features.
The results clearly show that using more accurate tattoo bounding boxes for compact feature learning does improve the tattoo matching accuracy.
However, we also notice that the CRL module achieves only $72\%$ rank-1 identification rate using the ground-truth tattoo bounding boxes.
The possible reason is that the small size of the training set (in terms of the number of tattoo classes and images) has limited the training of the CRL module.
There is still room to improve the performance of the CRL module.

\begin{figure*}[t]
\centering
\includegraphics[width=.83\linewidth,height=0.47\linewidth]{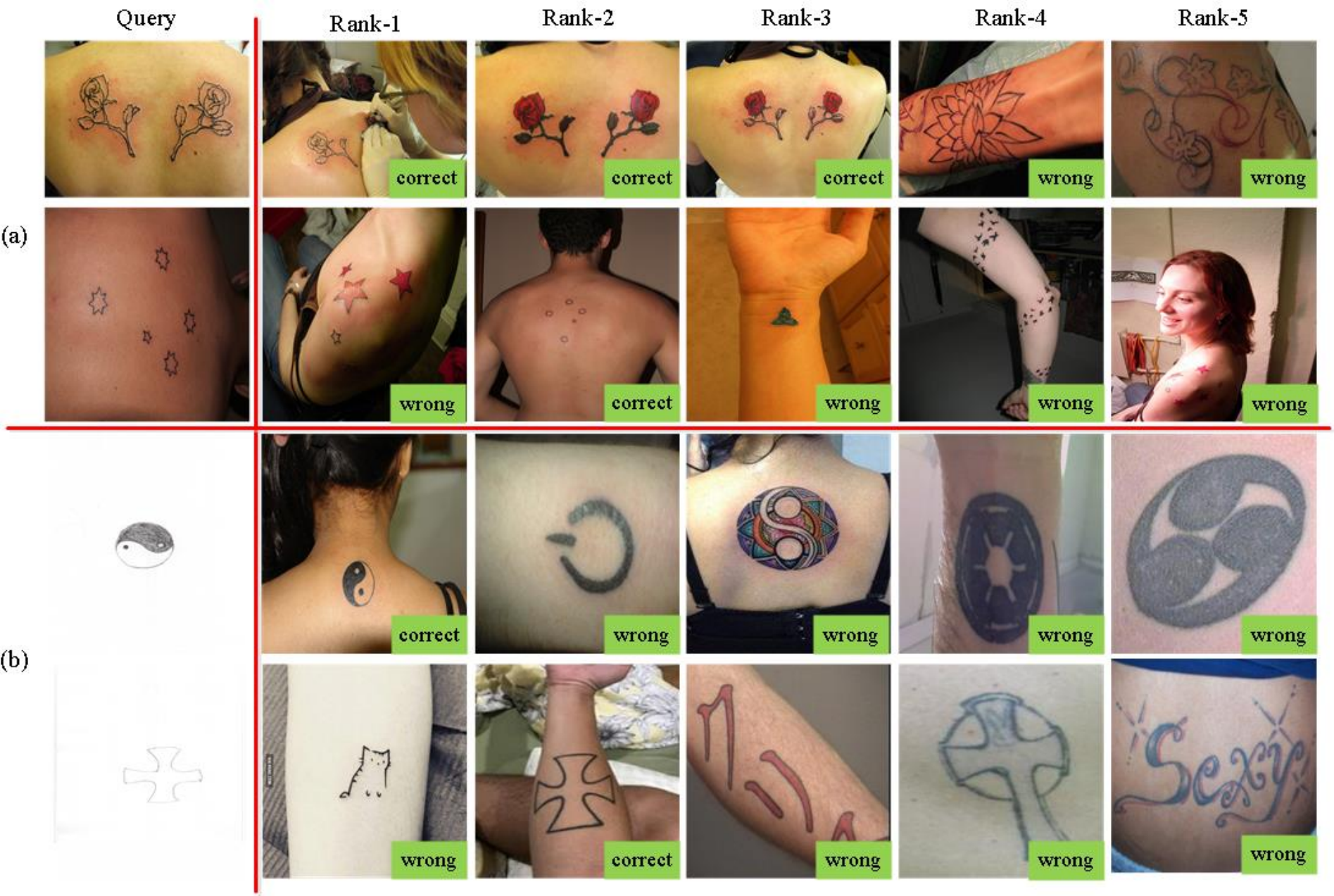}
\vspace{-3mm}
\caption[]{Examples of tattoo image search results by the proposed approach using (a) tattoo photos and (b) tattoo sketches as queries. For each query tattoo image, the top-5 tattoo gallery images in the returned list are given.}
\label{Fig.TattooSearchExamples}
\vspace{-4mm}
\end{figure*}

\begin{figure}[t]
\centering
\includegraphics[width=0.7\linewidth]{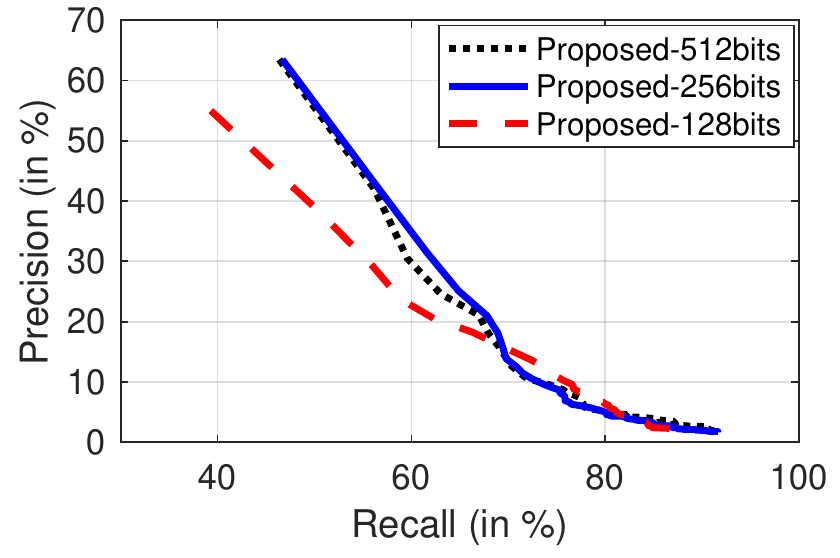}
\vspace{-4mm}
\caption[FeatDeimPR]{The influence of the compact binary code length (in bit) to the tattoo search performance (in terms of precision-recall) by the proposed approach reported on the WebTattoo test dataset without using background tattoo images in the gallery set.}
\label{Fig.FeatDimInfluence}
\vspace{-4mm}
\end{figure}

After we populate the gallery set using $300$K background tattoo images, as expected, all the approaches report decreased tattoo search performance (Fig. \ref{Fig.TattooSearchPR} (b)).
Matching the query tattoo images with the complete $300$K gallery using TattooID would take more than one week on CPU, so we use $100$K background tattoo images for TattooID.
Again, both OIM-ResNet50 and the proposed approach outperform TattooID and SSDH-VGG16 by a large margin, and our approach performs better than OIM-ResNet50.
This suggests that the proposed approach remains effective under large-scale tattoo search scenarios.

\begin{figure}[t]
\centering
\includegraphics[width=0.7\linewidth]{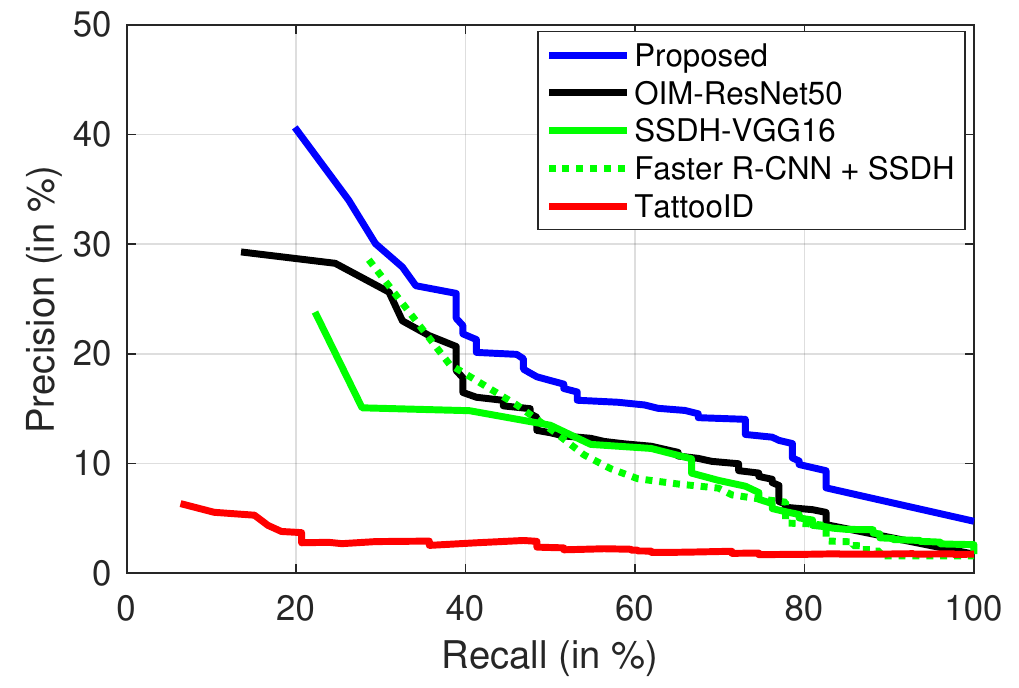}
\vspace{-4mm}
\caption[SketchSearchPR]{Sketch based tattoo search performance (in terms of precision-recall) by the proposed approach and the state-of-the-art methods (TattooID \cite{LeeMM12}, SSDH-VGG16 \cite{YangTPAMI17}, Faster R-CNN + SSDH, and OIM-ResNet50 \cite{XiaoCVPR17}) on the WebTattoo test dataset without background tattoo images in the gallery set.}
\label{Fig.SketchSearchPR}
\vspace{-4mm}
\end{figure}

We also evaluate the influence of different code lengths to the final tattoo search performance of our approach.
As shown in Fig. \ref{Fig.FeatDimInfluence}, when we increase the code length to $512$ bits, there is no performance improvement; instead, a minor precision drop is observed around $60\%$ recall.
If we reduce the code length to $128$ bits, there will be a large performance drop of the precision.
Therefore, we choose to use $256$-bit compact features in all the experiments of our approach.

Fig. \ref{Fig.TattooSearchExamples} (a) shows examples of tattoo search results by our approach on the WebTattoo database, in which the top-5 matched gallery images are given for each query tattoo image.
We can see that the proposed approach is robust against variations of body pose, illumination, and scale.
Some of the incorrectly matched gallery tattoo images by the proposed approach show high visual similarity to the query tattoo image (see the second row in Fig. \ref{Fig.TattooSearchExamples} (a)).

\subsection{Sketch Based Tattoo Search}
\label{Sec.SketchSearch}

In many scenarios, the surveillance image of the crime scene is not available, so the query is in the form of a sketch of a tattoo drawn based on the description provided by an eyewitness (see Fig. \ref{Fig.TattooSearchExamples} (b)). 
Therefore, it is important to evaluate the performance of a tattoo search system under a sketch based tattoo search scenario.
SSDH-VGG16 \cite{YangTPAMI17}, Faster R-CNN + SSDH, OIM-ResNet50 \cite{XiaoCVPR17}, and our method that are used in Sect. \ref{Sec.SearchatScale}, are fine-tuned on the tattoo sketch training set consisting of $240$ pairs of tattoo sketches and photos, and then evaluated on the tattoo sketch test set.

Fig. \ref{Fig.SketchSearchPR} shows the precision-recall curves of the proposed approach and the state-of-the-art methods for sketch-based tattoo search.
As expected, as a cross-modality search problem, tattoo sketch-to-photo matching is much more challenging than tattoo photo-to-photo matching.
Different from the observations in image-based tattoo search, SSDH-VGG16 performs better than TattooID in sketch-based tattoo search.
The main reasons are two-fold: (i) TattooID detects much less SIFT keypoints from the tattoo sketches drawn on the papers than from the tattoo photos; (ii) the learning based methods, such as SSDH-VGG16, are able to leverage the tattoo sketch-photo pairs to learn a feature representation that mitigates the modality gap between the tattoo sketches and photos.
The methods that compute features from the detected tattoos (\eg the proposed approach, OIM-ResNet50, and {Faster R-CNN + SSDH}) perform better than the methods that directly extract features from the holistic tattoo images.
The proposed approach performs consistently better than the state-of-the-art methods in sketch based tattoo search.
The results suggest that proposed joint tattoo detection and CRL approach has good generalization ability into the sketch-based tattoo search scenario.

Fig. \ref{Fig.TattooSearchExamples} (b) shows examples of sketch-based tattoo search results by our approach.
Benefited from the joint tattoo detection in the proposed approach, our feature representation can reduce the influence of the background regions of the tattoos in the gallery set, and thus is able to match a tattoo sketch to its mated tattoo image at a low rank.


\begin{figure}[t]
\centering
\includegraphics[width=0.63\linewidth]{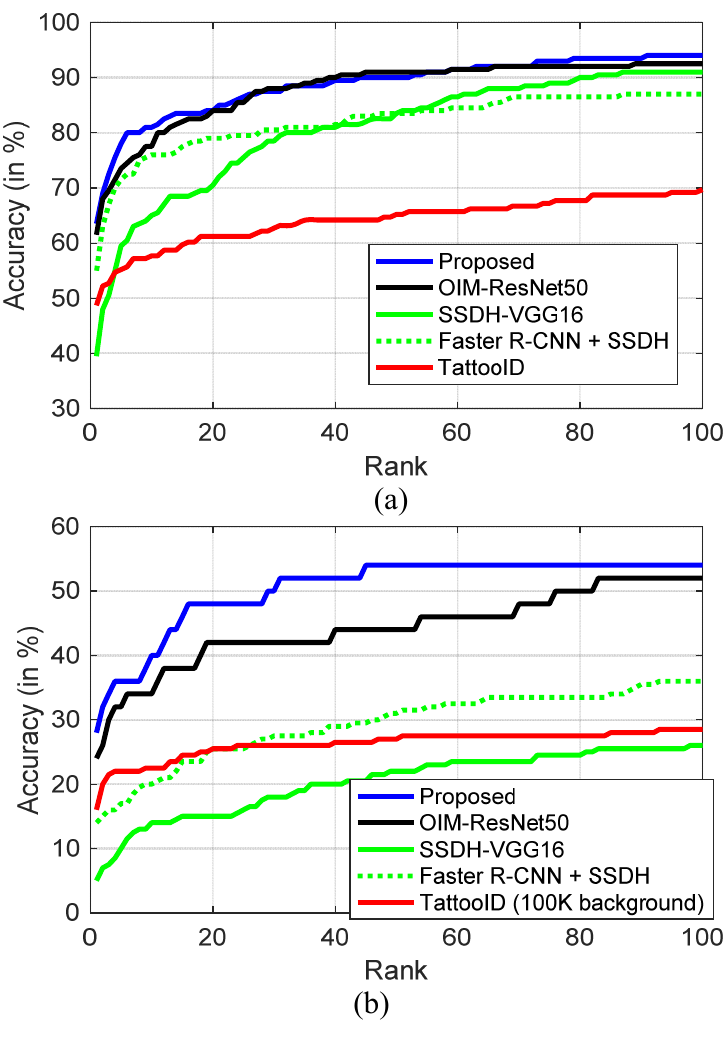}
\vspace{-4mm}
\caption[TattooSearchCMC]{Tattoo identification performance (in terms of CMC) by the proposed approach and the state-of-the-art methods (TattooID \cite{LeeMM12}, SSDH-VGG16 \cite{YangTPAMI17}, Faster R-CNN + SSDH, and OIM-ResNet50 \cite{XiaoCVPR17}) on the WebTattoo test dataset: (a) without $300$K background tattoo images in the gallery set, and (b) with $300$K background tattoo images in the gallery set; for TattooID, we report its tattoo search performance using an extended gallery set with only $100$K background images because of its long running time.}
\label{Fig.TattooSearchCMC}
\vspace{-4mm}
\end{figure}

\subsection{Tattoo Identification}
\label{Sec.TattooID}

Automatic tattoo identification techniques are usually utilized to generate a candidate suspect list, which is used for human or forensic analysis.
While high rank-1 accuracy is ideal, success in these forensic recognition scenarios is generally measured by the accuracies from rank-1 to rank-100 \cite{JainMultiMedia12}.
Therefore, a number of the published tattoo identification methods reported their performance in terms of CMC curves covering rank-1 to rank-100 \cite{Jain07,LeeMM12,HanICB13,ActonSSIAI08,XuICPR16}.
We report the CMC curves of the proposed approach and the state-of-the-art methods on the WebTattoo test dataset in Fig. \ref{Fig.TattooSearchCMC}.
Again, the results show that the methods that compute features from the detected tattoos (\eg the proposed approach, OIM-ResNet50, and Faster R-CNN + SSDH) perform better than the methods that directly extract features from the holistic tattoo images (\eg SSDH and TattooID).
However, the proposed approach, which leverages multi-task learning to perform joint tattoo detection and CFL, achieves the best accuracy.
When the $300$K background images are used to populate the gallery set, all the approaches are observed to have decreased identification accuracies, \eg about $30\%$ degradation at rank-1 identification accuracy.
The proposed approach still performs better than the baselines in such a challenging scenario.
These results indicate that the proposed approach has good generalization ability into the scenario of tattoo identification with a large gallery set.

\begin{figure}[t]
\centering
\includegraphics[width=0.63\linewidth]{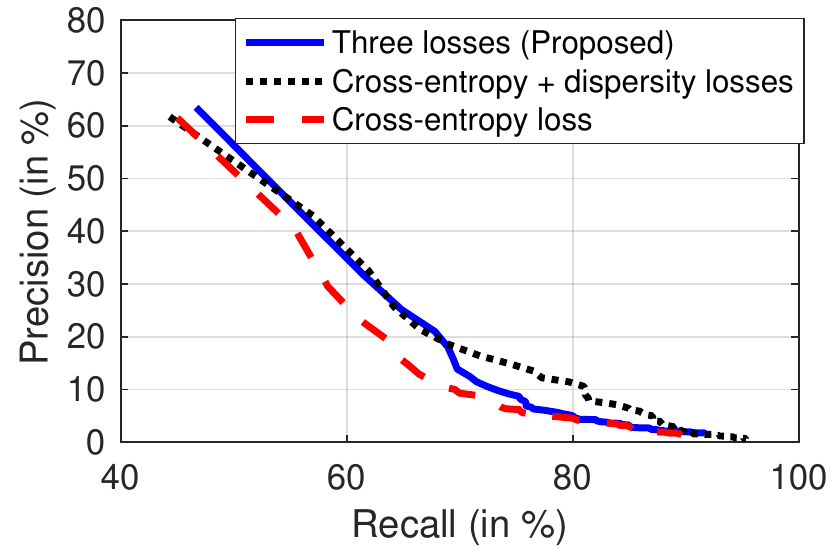}
\vspace{-4mm}
\caption[AblationStudyLosses]{Ablation studies involving three loss functions (cross-entropy, dispersity, and polarization) in CRL for tattoo image search on the WebTattoo test dataset without distracter images in the gallery set.}
\label{Fig.AblationStudyLosses}
\vspace{-4mm}
\end{figure}

On the public Tatt-C identification dataset, the proposed approach achieves $99.2\%$ rank-1 identification accuracy.
The MorphoTrak and the Purdue teams reported $99.4\%$ and $98.7\%$ rank-1 identification accuracies on the Tatt-C identification dataset.
While the results by MorphoTrak are slightly better than ours, they used four folds of data of Tatt-C for training, and the fifth-fold data for testing.
By contrast, our approach is trained on the WebTattoo training dataset, which is different from the tattoo images in Tatt-C.

In addition to the above evaluations, we also evaluate the generalization ability of the proposed approach in other instance-level retrieval tasks, such as on Paris \cite{PhilbinCVPR08} and Oxford \cite{PhilbinCVPR07}, which contain challenging viewpoint and scale variations.
Following the same testing protocol as the state-of-the-art method \cite{YangTPAMI17}, our approach achieves $83.24\%$ and $53.04\%$ mAP on Paris and Oxford, respectively, which are comparable to the performance of state-of-the-art method \cite{YangTPAMI17} ($83.87\%$ and $63.79\%$ mAP on Paris and Oxford, respectively).
These results show that the proposed approach has good generalization ability to new application scenarios.

\subsection{Ablation Study}
\label{Sec.AblationStudy}

We provide ablation studies of our approach in terms of three loss functions, \ie (i) basic cross-entropy loss, (ii) cross-entropy loss and dispersity loss, and (ii) all three losses together.
The precision-recall curves calculated using the top-100 retrieval results of the three experiments are shown in Fig. \ref{Fig.AblationStudyLosses}.
We can see that using dispersity loss together with the cross-entropy loss does not improve the rank-1 tattoo search accuracy compared to using cross-entropy loss alone, but it does improve the overall performance beyond rank-1.
Jointly using all three losses leads to the best performance, particularly for the rank-1 tattoo search accuracy; this is important for practical applications.
The reason why jointly using all three losses works better for compact feature learning is that while cross-entropy loss is helpful for generating real-valued codes that are discriminative between individual tattoo classes, dispersity and polarization losses assure the real-valued codes are near-binary and evenly distributed in the code space (see our explanations in Sect. \ref{Sec.JointDet&Feature} and Fig. \ref{Fig.PolarizationLoss}).

\subsection{Computational Cost}
\label{Sec.ComputationalCost}

We summarize the computational cost of the proposed approach and several state-of-the-art methods.
Our approach takes about $0.2$ sec. in total to perform joint detection and CRL on a Titan X GPU.
After obtaining the compact feature representation (256-bit), the average time of computing the Hamming distance of two 256-D binary codes is $0.06$ms on an Intel i7 $3.6$GHz CPU without using bitwise operation based optimizations, which is $5$ times faster than computing the cosine distance of two 256-D real-valued codes ($0.3$ms on average).
Such a difference in computational cost matters particularly for scenarios of tattoo search from huge volumes of surveillance video frames or handling multiple parallel searching requests.
At the same time, comparisons with the state-of-the-art methods based on real-valued codes, \eg \cite{XiaoCVPR17}, show that our compact binary codes of the same length can achieve better accuracy.
Only a few of the published tattoo identification and tattoo retrieval methods have reported their computational costs.
For example, \cite{JainICIP09} reported an average of $24$ sec. in comparing one query tattoo against $10$K gallery tattoos after obtaining the SIFT features on an Intel Core2 $2.66$ GHz CPU, which is much slower than the proposed compact representation.
We also profiled the tattoo detection time by a Faster R-CNN detector in \cite{SunICPR16} (using the same input image size as our approach) and the feature extraction time by a VGG-16 CNN network used in \cite{YangTPAMI17}.
Tattoo detection and feature extraction per tattoo detection take $0.2$ sec. and $0.04$ sec., respectively.
This indicates that the proposed approach is more efficient in real application scenarios, in which there are usually more than one tattoo detections per image.

\section{Conclusions}
\label{Sec.Summary}

This paper presents a joint detection and compact feature learning approach for tattoo image search at scale.
While existing tattoo search methods mainly focus on matching cropped tattoos, the proposed approach models tattoo detection and compact representation learning in a single convolutional neural network via multi-task learning.
The WebTattoo dataset consisting of $300$K tattoo images was compiled from the public-domain tattoo datasets and images from the Internet.
In addition, $300$ tattoo sketches were created for sketch-based tattoo search to replicate the scenario where the surveillance image of the tattoo is not available.
These datasets help evaluate the proposed approach for tattoo image search at scale and in operational scenarios.
Our approach performs well on a number of tasks including tattoo detection, tattoo search at scale, and sketch-based tattoo search.
The proposed data augmentation method is able to replicate various tattoo appearance variations, and thus is helpful to improve the robustness of the tattoo detector in unconstrained scenarios.
Experimental results with cross-database testing protocols show that the proposed approach generalizes well to the unseen scenarios.


%



\ifCLASSOPTIONcompsoc
  \section*{Acknowledgments}
\else
  \section*{Acknowledgment}
\fi

This research was supported in part by the Natural Science Foundation of China (grants 61732004, 61390511, and 61672496), External Cooperation Program of Chinese Academy of Sciences (CAS) (grant GJHZ1843), and Youth Innovation Promotion Association CAS (2018135).
﻿The preliminary work appeared in the Proceedings of the 6th International Conference on Biometrics (ICB), 2013 \cite{HanICB13}.
Anil K. Jain is the corresponding author.

\ifCLASSOPTIONcaptionsoff
  \newpage
\fi



\bibliographystyle{IEEEtran}
\bibliography{references}
%
%
%

%


\begin{IEEEbiography}[{\includegraphics[width=1in,height=1.25in,clip,keepaspectratio]{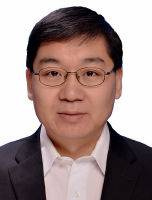}}]{Hu Han} is an Associate Professor of the Institute of Computing Technology (ICT), Chinese Academy of Sciences (CAS).
He received the B.S. degree from Shandong University, and the Ph.D. degree from ICT, CAS,  in 2005 and 2011, respectively, both in computer science.
Before joining the faculty at ICT, CAS in 2015, he has been a Research Associate at PRIP lab in the Department of Computer Science and Engineering at Michigan State University, and a Visiting Researcher at Google in Mountain View.
His research interests include computer vision, pattern recognition, and image processing, with applications to biometrics, forensics, law enforcement, and security systems. He is a member of the IEEE.
\end{IEEEbiography}
\begin{IEEEbiography}[{\includegraphics[width=1in,height=1.25in,clip,keepaspectratio]{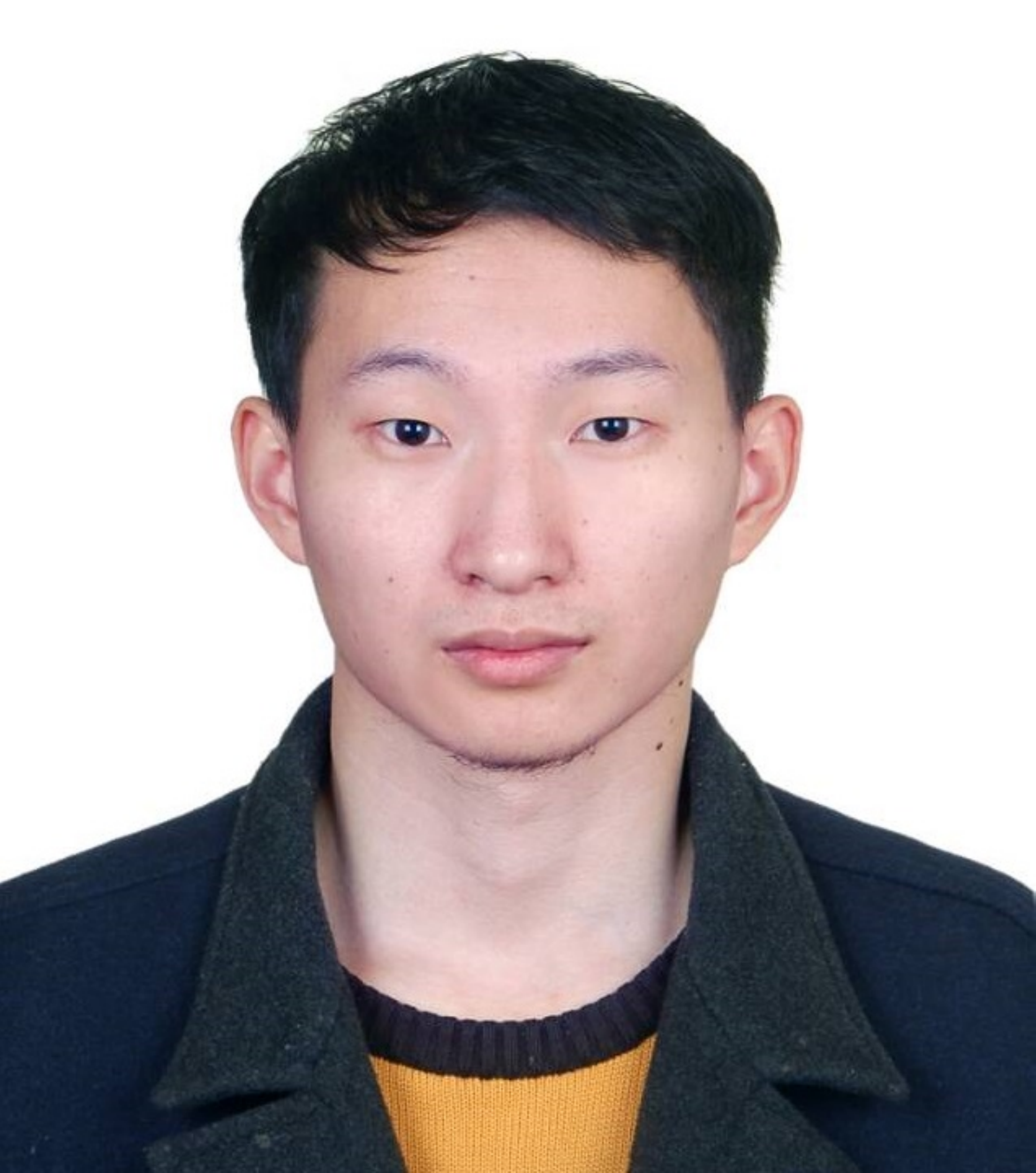}}]{Jie Li} received the B.S. degree from Qingdao University in 2017, and he is working toward the M.S. degree in ICT, CAS and the University of Chinese Academy of Sciences. His research interests include computer vision, pattern recognition, and image processing, with applications to biometrics.
\end{IEEEbiography}

\begin{IEEEbiography}[{\includegraphics[width=1in,height=1.25in,clip,keepaspectratio]{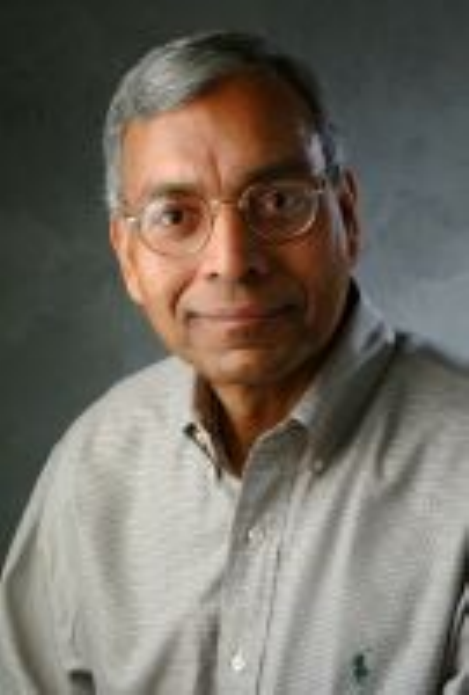}}]{Anil K. Jain} is a University Distinguished Professor in the Department of Computer Science and Engineering at Michigan State University. His research interests include pattern recognition and biometric authentication. He served as the editor-in-chief of the \textsc{IEEE Transactions on Pattern Analysis and Machine Intelligence} (1991-1994). He served as a member of the United States Defense Science Board and The National Academies committees on Whither Biometrics and Improvised Explosive Devices. He has received Fulbright, Guggenheim, Alexander von Humboldt, and IAPR King Sun Fu awards. He is a member of the National Academy of Engineering
and foreign fellow of the Indian National Academy of Engineering. He is a Fellow of the AAAS, ACM, IAPR, SPIE, and IEEE.
\end{IEEEbiography}

\begin{IEEEbiography}[{\includegraphics[width=1in,height=1.25in,clip,keepaspectratio]{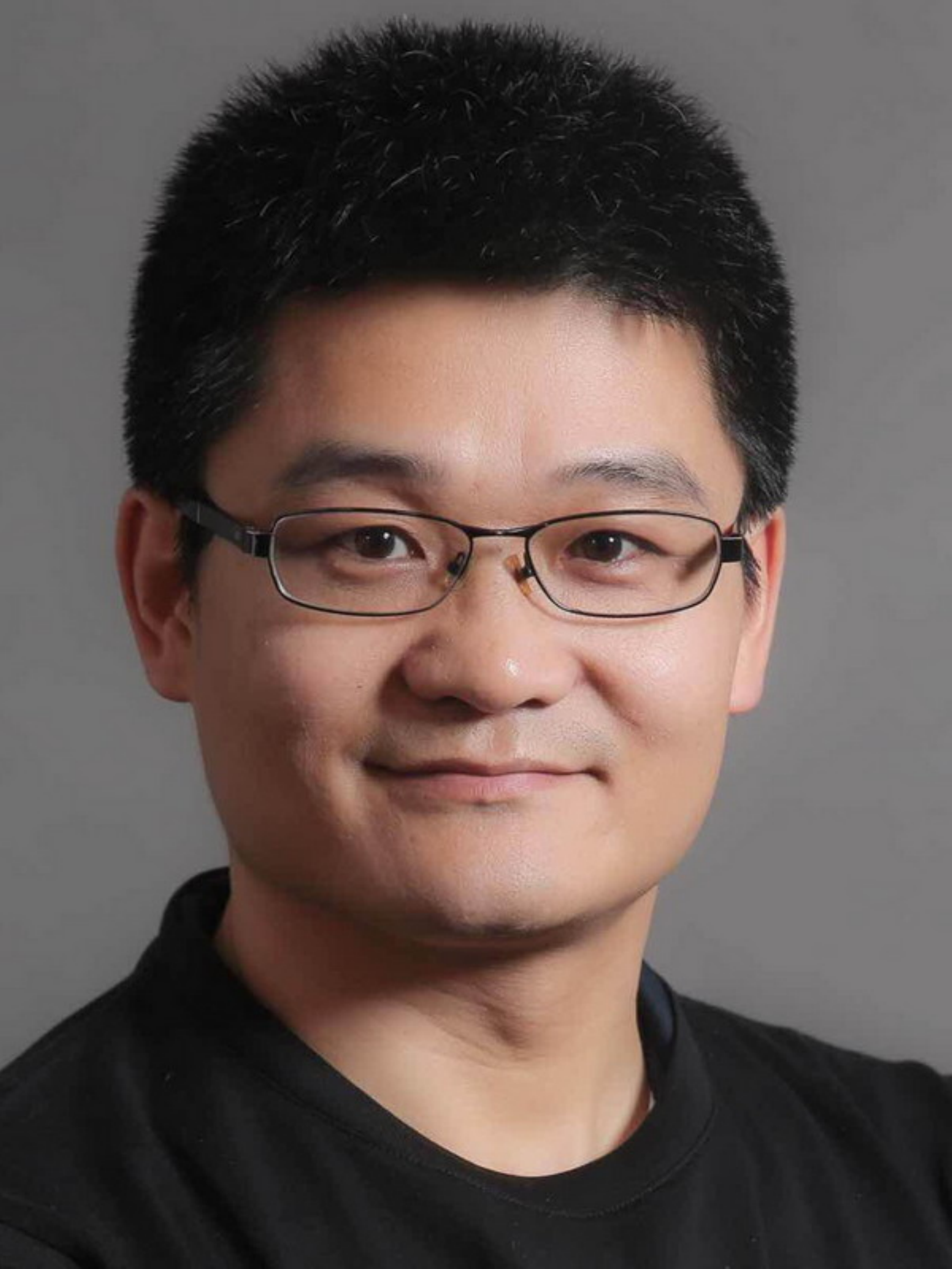}}]{Shiguang Shan} is a Professor of
ICT, CAS, and the Deputy Director with the Key Laboratory of Intelligent Information Processing, CAS.
His research interests cover computer vision, pattern recognition, and machine learning.
He has authored over 200 papers in refereed journals and proceedings in the areas of computer vision and pattern recognition.
He was a recipient of the China's State Natural Science Award in 2015, and the China's State S\&T Progress Award in 2005 for his research work.
He has served as the Area Chair for many international conferences, including ICCV'11, ICPR'12, ACCV'12, FG'13, ICPR'14, and ACCV'16. He is an Associate Editor of several journals, including the \textsc{IEEE Transactions on Image Processing}, the Computer Vision and Image Understanding, the Neurocomputing, and the Pattern Recognition Letters.
He is a Senior Member of IEEE.
\end{IEEEbiography}

\begin{IEEEbiography}[{\includegraphics[width=1in,height=1.25in,clip,keepaspectratio]{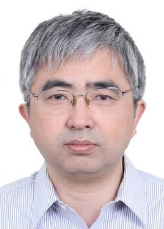}}]{Xilin Chen} is a Professor of
ICT, CAS.
He has authored one book and over 200 papers in refereed journals and proceedings in the areas of computer vision, pattern recognition, image processing, and multimodal interfaces.
He served as an Organizing Committee/Program Committee member for over 70 conferences. He was a recipient of several awards, including the China's State Natural Science Award in 2015, the China's State S\&T Progress Award
in 2000, 2003, 2005, and 2012 for his research work. He is currently an Associate Editor of the \textsc{IEEE Transactions on Multimedia}, a Leading Editor of the Journal of Computer Science and Technology, and an Associate Editor-in-Chief of the Chinese Journal of Computers.
He is a Fellow of the China Computer Federation (CCF), IAPR, and IEEE.
\end{IEEEbiography}



\end{document}